\PassOptionsToPackage{numbers}{natbib}
\documentclass{article}

\usepackage{natbib}
\usepackage[margin=1in]{geometry}
\usepackage[utf8]{inputenc}
\usepackage[T1]{fontenc}
\usepackage{hyperref}
\usepackage{url}
\usepackage[ruled,vlined,algo2e]{algorithm2e}
\usepackage{booktabs}
\usepackage{amsfonts}
\usepackage{nicefrac}
\usepackage{microtype}
\usepackage{xcolor}
\usepackage{mathtools}
\usepackage{float}
\usepackage{amsmath,amsfonts,amssymb,amsthm}
\usepackage{thmtools,thm-restate}
\usepackage{bm}
\usepackage{enumitem,array}
\usepackage{algorithmic}
\usepackage{algorithm}
\usepackage{bbm}
\usepackage{tabulary,multirow}
\usepackage{dsfont}
\usepackage{thmtools,thm-restate}
\usepackage{multicol}
\usepackage{authblk}
\usepackage{upgreek}
\usepackage{tabularx}
\usepackage{lipsum}
\usepackage{enumitem}
\usepackage{authblk}
\date{}

\newcommand{\detopt}{\mathrm{OPT}}
\newcommand{\minactions}{\cA_-}
\newcommand{\maxactions}{\cA_+}

\newcommand{\hhyp}{f}
\newcommand{\minop}{\minplayerx{g}}
\newcommand{\maxop}{\maxplayerx{g}}
\newcommand{\vierr}{\mathrm{VErr}}
\newcommand{\vcdt}{\mathrm{VC}}
\newcommand{\lsdt}{\mathrm{LD}}

\newcommand{\lipschitz}{L}

\newcommand{\equationref}[1]{Equation #1}

\newtheorem{theorem}{Theorem}[section]
\newtheorem{lemma}[theorem]{Lemma}
\newtheorem{definition}{Definition}[section]

\newtheorem{corollary}[theorem]{Corollary}

\newtheorem{fact}{Fact}[section]

\newtheorem{claim}{Claim}[section]

\renewcommand{\epsilon}{\varepsilon}
\renewcommand{\hat}{\widehat}
\renewcommand{\tilde}{\widetilde}

\newcommand{\innerproduct}[2]{\left< #1, #2 \right>}
\newcommand{\inner}[2]{\left< #1, #2 \right>}

\newcommand{\norm}[1]{\left\lVert#1\right\rVert}

\newcommand{\para}[1]{\left(#1\right)}

\newcommand{\paraf}[1]{(#1)}
\newcommand{\parantheses}[1]{\left(#1\right)}

\newcommand{\curlybrackets}[1]{\left\{#1\right\}}

\newcommand{\bset}[1]{\curlybrackets{#1}}

\newcommand{\bsetf}[1]{\{#1\}}

\newcommand{\abs}[1]{\left|#1\right|}

\newcommand{\setsize}[1]{\left| #1 \right|}

\DeclareMathOperator*{\Exp}{\mathbb{E}}

\newcommand{\EE}[1]{\Exp\left[#1\right]}

\newcommand{\EEs}[2]{\Exp_{#1}\left[#2\right]}

\newcommand{\EEsf}[2]{\Exp_{#1}[#2]}

\newcommand{\EEc}[2]{\Exp\left[#1\left|#2\right.\right]}

\renewcommand{\cite}[1]{\citep{#1}}

\newcommand{\bigO}[1]{O\parantheses{#1}}

\newcommand{\reals}{\mathbb{R}}

\newcommand{\integers}{\mathbb{Z}}

\newcommand{\simplex}{\Delta}
\newcommand{\simiid}{\mathrel{\stackrel{\makebox[0pt]{\mbox{\normalfont\tiny i.i.d.}}}{\sim}}}

\DeclareMathOperator*{\argmin}{arg\,min}

\newcommand{\asseq}{\coloneqq}

\newcommand{\datapoints}{\cZ}

\newcommand{\features}{\mathcal{X}}

\newcommand{\labels}{\mathcal{Y}}

\newcommand{\hedgealgorithm}{\mathrm{Hedge}}

\newcommand{\learningalgorithm}{\cQ}
\newcommand{\exoracle}{\text{EX}}

\newcommand{\vcd}{d}

\newcommand{\risk}{\cR}
\newcommand{\loss}{\ell}
\newcommand{\losses}{\cL}
\newcommand{\numlosses}{m}

\newcommand{\opt}{\mathrm{OPT}}
\newcommand{\regret}{\mathrm{Reg}}

\newcommand{\hyp}{h}

\newcommand{\hyps}{\cH}

\newcommand{\dist}{D}

\newcommand{\dists}{\cD}

\newcommand{\numdists}{n}

\newcommand{\params}{\Theta}

\newcommand{\convexset}{Z}
\newcommand{\convexsetcore}{\Theta^o}

\newcommand{\dgf}{\omega}
\newcommand{\proxfunction}{V}

\newcommand{\stepsize}{\eta}

\newcommand{\diameter}{R}

\newcommand{\probinstance}{\mathbb{P}}
\newcommand{\instanceset}{\mathbb{V}}

\newcommand{\instance}{V}

\newcommand{\timestep}{t}

\newcommand{\tsv}[2]{#1\vphantom{#1}^{\parantheses{#2}}}

\newcommand{\payoff}{\phi}

\newcommand{\cost}{{c}}

\newcommand{\action}{a}

\newcommand{\actions}{A}
\newcommand{\numactions}{n}

\newcommand{\minplayerx}[1]{#1_\text{-}}
\newcommand{\maxplayerx}[1]{#1_\text{+}}
\newcommand{\minx}[1]{#1_\text{-}}
\newcommand{\maxx}[1]{#1_\text{+}}

\newcommand{\minaction}{p}
\newcommand{\maxaction}{q}
\newcommand{\mina}{\minaction}
\newcommand{\maxa}{\maxaction}

\newcommand{\maxweight}{q}

\newcommand{\cA}{\mathcal{A}}

\newcommand{\cD}{\mathcal{D}}

\newcommand{\cF}{\mathcal{F}}

\newcommand{\cH}{\mathcal{H}}

\newcommand{\cL}{\mathcal{L}}

\newcommand{\cP}{\mathcal{P}}
\newcommand{\cQ}{\mathcal{Q}}
\newcommand{\cR}{\mathcal{R}}

\newcommand{\cX}{\mathcal{X}}

\newcommand{\cZ}{\mathcal{Z}}

\newcommand{\bx}{{\mathbf{x}}}

\title{
On-Demand Sampling: Learning Optimally from Multiple Distributions%
\thanks{Authors are ordered alphabetically. Correspondence to \texttt{eric.zh@berkeley.edu}.}
}

\author[]{Nika Haghtalab}
\author[]{Michael I. Jordan}
\author[]{Eric Zhao}
\affil[]{University of California, Berkeley\\{ \texttt{\{nika,jordan,eric.zh\}@berkeley.edu}}}

\begin{document}
\allowdisplaybreaks
\maketitle

\begin{abstract}
    Social and real-world considerations such as robustness, fairness, social welfare and multi-agent tradeoffs have given rise to multi-distribution learning paradigms, such as \emph{collaborative}~\citep{blum_collaborative_2017}, \emph{group distributionally robust} ~\citep{sagawa_distributionally_2020}, and \emph{fair federated} learning~\citep{mohri_agnostic_2019}.
	In each of these settings, a learner seeks to uniformly minimize its expected loss over $\numdists$ predefined data distributions, while using as few samples as possible.
	In this paper, we establish the optimal sample complexity of these learning paradigms and give algorithms that meet this sample complexity.
	Importantly, our sample complexity bounds exceed that of learning a single distribution by only an additive factor of $\frac{\numdists \log(\numdists)}{\epsilon^2}$.
	This improves upon the best known sample complexity bounds for fair federated learning (by \citet{mohri_agnostic_2019}) and collaborative learning (by \citet{nguyen_improved_2018}) by multiplicative factors of $\numdists$ and $\frac{\log(\numdists)}{\epsilon^3}$, respectively.
	We also provide the first sample complexity bounds for the \emph{group DRO} objective of \citet{sagawa_distributionally_2020}.
	To guarantee these optimal sample complexity bounds, our algorithms learn to sample from data distributions on demand.
	Our algorithm design and analysis are enabled by our extensions of online learning techniques for solving stochastic zero-sum games.
	In particular, we contribute stochastic variants of no-regret dynamics that can trade off between players' differing sampling costs.
\end{abstract}

\section{Introduction}
Pervasive needs for robustness, fairness, and multi-agent collaboration in learning have given rise to multi-distribution learning paradigms (e.g., \citep{blum_collaborative_2017,sagawa_distributionally_2020,mohri_agnostic_2019, duchi_learning_2018}).
In these settings, we seek to learn a model that performs well on \emph{any distribution} in a predefined set of interest.
For fairness considerations, these distributions may represent heterogeneous populations of different protected or socioeconomic attributes; in robustness applications, they may capture a learner's uncertainty regarding the true underlying task; and in multi-agent collaborative or federated applications, they may represent agent-specific learning tasks.
In these applications, the performance and optimality of a model is measured by its worst test-time performance on a distribution in the set.
We are concerned with this fundamental problem of designing sample-efficient multi-distribution learning algorithms.

The sample complexity of multi-distribution learning differs from that of learning a single distribution in several ways.
On one hand, varying numbers of samples are required when learning tasks of varying difficulty.
On the other hand, similarity or overlap among learning tasks may obviate the need to sample from some distributions.
This makes the use of a fixed per-distribution sample budget highly inefficient and suggests that optimal multi-distribution learning algorithms should \emph{sample on demand}.
That is, algorithms should take additional samples \emph{whenever they need them} and \emph{from whichever data distribution} they want them.
On-demand sampling is especially appropriate when some population data is scarce (as in fairness mechanisms in which samples are amended~\cite{ramaswamy_fair_2021}); when the designer can actively perturb datasets towards rare or atypical instances (such as in robustness applications~\cite{kar_meta-sim_2019, zakharov_deceptionnet_2019}); or when sample sets represent agents' contributions to an interactive multi-agent system~\cite{mohri_agnostic_2019, blum_one_2021}.

\citet{blum_collaborative_2017} demonstrated the benefit of on-demand sampling in the \emph{collaborative learning} setting, when all data distributions are realizable with respect to the same target classifier.
This line of work established that learning $\numdists$ distributions with on-demand sampling requires a factor of $\tilde{O}(\log(\numdists))$ times the sample complexity of learning a single realizable distribution~\cite{blum_collaborative_2017,chen_tight_2018,nguyen_improved_2018}, whereas relying on batched uniform convergence takes $\tilde{\Omega}(\numdists)$ times more samples than learning a single distribution~\cite{blum_collaborative_2017}.
However, beyond the realizable setting, the best known multi-distribution learning results fall short of this promise: existing on-demand sample complexity bounds for agnostic collaborative learning have highly suboptimal dependence on $\epsilon$, requiring $\tilde{O}(\log(\numdists)/\epsilon^3)$ times the sample complexity of agnostically learning a single distribution~\cite{nguyen_improved_2018}.
On the other hand, agnostic fair federated learning bounds~\cite{mohri_agnostic_2019} have been studied only for algorithms that sample in one large batch and thus require $\tilde{\Omega}(\numdists)$ times the sample complexity of a single learning task.
Moreover, the test-time performance of some key multi-distribution learning methods, such as group distributionally robust optimization~\cite{sagawa_distributionally_2020}, have not been studied from a provable or mathematical perspective before.

\renewcommand{\arraystretch}{2}
\begin{table}
	\centering
	\small{
		\begin{tabular*}{\textwidth}{l l c l}
			\hline
			Problem & Sample Complexity & Thm & Best Previous Result \\
			\hline
			Collab. Learning UB & $\epsilon^{-2} \paraf{\log(\setsize{\hyps}) + \numdists \log (\numdists/\delta) }$ & [\ref{thm:finite}] & $ \epsilon^{-5} \log\paraf{\frac{1}{\epsilon}} \log(\numdists/\delta) (\log(\setsize{\hyps}) + \numdists)$~\cite{nguyen_improved_2018} \\
			Collab. Learning LB & $\epsilon^{-2}(\log(\setsize{\hyps}) + \numdists \log (\numdists/\delta))$ & [\ref{theorem:lowerboundagnostic}] & $\epsilon^{-1} (\log(\setsize{\hyps}) + \numdists \log (\numdists / \delta))$~\cite{blum_collaborative_2017} \\
			\hline
			GDRO/AFL UB & $\epsilon^{-2} \paraf{\log(\setsize{\hyps}) + \numdists \log(\numdists/\delta)}$& [\ref{thm:finite}] & $\epsilon^{-2} \paraf{n \log(\setsize{\hyps}) + \numdists \log(\numdists/\delta)}$~\cite{mohri_agnostic_2019} \\
			GDRO/AFL UB & $\epsilon^{-2} \paraf{\dist_{\hyps} + \numdists \log(\numdists/\delta)}$ & [\ref{thm:gdro}] & N/A \\
			(Training error convg.) & $\epsilon^{-2} \paraf{\dist_{\hyps} + \numdists \log(\numdists/\delta)}$ & [\ref{corr:gdro}] & $\epsilon^{-2} \numdists (\log(\numdists) + D_{\hyps})$ (expected convergence only)~\cite{sagawa_distributionally_2020} \\
			\hline
		\end{tabular*}
	}
	\caption{
		\small
		This table lists upper (\textit{UB}) and lower bounds (\textit{LB}) on the sample complexity of learning a model class $\hyps$ on $\numdists$ distributions.
		For the collaborative learning and agnostic federated learning (AFL) settings, the sample complexity upper bounds refer to the problem of learning a (potentially randomized) model whose expected loss on each distribution is at most $\opt + \epsilon$, where $\opt$ is the best possible such guarantee.
		For the GDRO setting, sample complexity refers to learning a deterministic model with expected losses of at most $\opt + \epsilon$, from a convex compact model space $\hyps$ with a Bregman radius of $\dist_\hyps$.
		Sample complexity bounds for collaborative and agnostic federated learning in existing works extend to VC dimension and Rademacher complexity.
		Our results also extend to VC dimension under some assumptions.
	}
	\label{tab:summary}
\end{table}

In this paper, we give a general framework for obtaining \emph{optimal and on-demand sample complexity} for three multi-distribution learning settings.
Table~\ref{tab:summary} summarizes our results.
All three of these settings consider a set $\dists$ of $\numdists$ data distributions and a model class $\hyps$, evaluating the performance of a model $\hyp$ by its worst-case expected loss, $\max_{\dist \in\dists} \risk_\dist(\hyp)$.
As a benchmark, they consider the worst-case expected loss of the best model, i.e., $\detopt = \min_{\hyp^*\in\hyps} \max_{\dist \in \dists} \risk_\dist(\hyp^*)$.
Notably, all of our sample complexity upper bounds demonstrate only an \emph{additive increase of $\epsilon^{-2} \numdists \log(\numdists / \delta)$ over the sample complexity of a single learning task}, compared to the multiplicative factor increase required by existing works.
\begin{itemize}[ leftmargin=10pt]
	\item[-] \emph{Collaborative learning of~\citet{blum_collaborative_2017}:}
		For agnostic collaborative learning, our Theorem~\ref{thm:finite} gives a randomized and a deterministic model that achieves performance guarantees of $\detopt+\epsilon$ and $2 \detopt + \epsilon$, respectively.
		Our algorithms have an optimal sample complexity of $O\paraf{\frac{1}{\epsilon^2}(\log(\setsize{\hyps}) + \numdists \log(\numdists/\delta))}$.
		This improves upon the work of~\citet{nguyen_improved_2018} in two ways.
		First, it provides risk bounds of $\detopt+\epsilon$ for randomized classifiers, where only $2 \detopt+\epsilon$ was established previously.
		Second, it improves the upper bound of~\citet{nguyen_improved_2018} by a multiplicative factor of $\log(\numdists) / \epsilon^3$.
		In Theorem~\ref{theorem:lowerboundagnostic}, we give a matching lower bound on this sample complexity, thereby establishing the optimality of our algorithms.

	\item[-] \emph{Group distributionally robust learning (group DRO) of~\citet{sagawa_distributionally_2020}:}
		For group DRO, we consider a convex and compact model space $\hyps$.
		Our Theorem~\ref{thm:gdro} studies a model that achieves an $\detopt+\epsilon$ guarantee on the worst-case test-time performance of the model with an on-demand sample complexity of $\bigO{\frac{1}{\epsilon^2} (D_\hyps + \numdists \log (\numdists / \delta))}$.
		Our results also imply a high-probability bound for the convergence of group DRO training error that improves upon the (expected) convergence guarantees of~\citet{sagawa_distributionally_2020} by a factor of $\numdists$.

	\item[-] \emph{Agnostic federated learning of ~\cite{mohri_agnostic_2019}:}
		For agnostic federated learning, we consider a finite class of hypotheses.
		Our Theorems~\ref{thm:finite} and \ref{thm:gdro} show that on-demand sampling can accelerate the generalization of agnostic federated learning by a factor of $\numdists$ compared to batch results established by~\citet{mohri_agnostic_2019}.
		Our results also imply matching high-probability bounds with respect to~\citet{mohri_agnostic_2019} on the convergence of the training error in the batched setting.
\end{itemize}

To achieve these results, we frame multi-distribution learning as a stochastic zero-sum game: a maximizing player chooses a weight vector over data distributions $\dists$ and a minimizing player chooses a weight vector over hypotheses $\hyps$.
These two players require different numbers of datapoints in order to estimate their respective payoff vectors.
We therefore solve the game using no-regret dynamics, utilizing stochastic mirror descent to optimally trade off the players' asymmetric needs for datapoints.
In Section~\ref{sec:tech-overview}, we give an overview of this approach and its technical challenges and contributions.
Our results also extend directly to settings with not only multiple data distributions but also multiple loss functions.

\subsection{Related Work}
There are many lines of work that study multi-distribution learning but which have evolved independently in separate communities.

\paragraph{Collaborative and agnostic federated learning.}
\citet*{blum_collaborative_2017} posed the first fully general description of multi-distribution learning, motivated by the application of \emph{collaborative PAC learning}.
The field of \textit{collaborative learning} is concerned with the learning of a shared machine learning model by multiple \textit{stakeholders} that each desire a model with low error on their own data distribution.
The line of work studies on-demand sample complexity bounds for the setting where stakeholders collect data so as to minimize the error of the worst-off stakeholder~\cite{blum_collaborative_2017, nguyen_improved_2018, chen_tight_2018,blum_communication-aware_2021}.
This setting, stated in its full generality, yields the multi-distribution learning problem as presented in this paper.
\citet{blum_collaborative_2017} established a $\log(\numdists)$ factor blowup for the realizable case. For the general agnostic setting the best existing sample complexity requires a factor $\log(\numdists) / \epsilon^3$ blowup~\cite{nguyen_improved_2018}.
In comparison, our work establishes a tight additive increase in the sample complexity (which is comparable to $\log(\numdists)$ multiplicative factor blowup with no dependence on $\epsilon$).
A related line of work concerns the strategic considerations of collaborative learning and seeks incentive-aware mechanisms for collecting data in the collaborative learning setting~\cite{blum_one_2021}.

The field of \textit{federated learning} focuses on a related motivating application where the goal is to learn a model from data dispersed across multiple devices but where querying data from each device is expensive~\cite{mcmahan_communication-efficient_2017}.
The \emph{agnostic} federated learning framework of~\citet*{mohri_agnostic_2019} poses (a variant of) the multi-distribution learning objective as a target for federated learning algorithms, and studies it in the offline setting with a data-dependent analysis.
Their results involve a blowup by a factor $\numdists$ for the sample complexity.

\paragraph{Group distributionally robust optimization (Group DRO).}
Multi-distribution learning also arises in distributionally robust optimization~\cite{ben-tal_robust_2009} under the name of Group DRO, a class of DRO problems where the distributional uncertainty set is finite~\cite{hashimoto_fairness_2018}.
The group DRO literature is motivated by applications where the distributions correspond to deployment domains or protected demographics that a machine learning model should avoid spuriously linking to labels~\cite{hashimoto_fairness_2018, sagawa_distributionally_2020, sagawa_investigation_2020}.
Although Group DRO---like collaborative learning---is mathematically an instance of multi-distribution learning, prior work on Group DRO focuses on the convergence of training error in offline settings, with a particular focus on deep learning applications.
As we discuss later, theoretical aspects of on-demand multi-distribution learning can translate into actionable insights for Group DRO applications.

\paragraph{Multi-group fairness.}
Multi-distribution learning is also related to the fields of multi-group learning~\cite{rothblum_multi-group_2021, tosh_simple_2022} and multi-group fairness~\cite{dwork_outcome_2021, hu_metric_2022}.
These works study offline learning settings with a single distribution $\dist$ and implicitly consider distribution $\dist_i$ to be the conditional distribution on a subset of the support representing group $i$.
In these settings, the learner does not have explicit access to oracles that sample from distributions $\dist_1, \dots, \dist_\numdists$ and instead uses rejection sampling to collect data from $\dist_1, \dots, \dist_\numdists$.
As a result, they experience a sub-optimal sample complexity blowup by a factor $\numdists$.
This blowup may not be obvious upon first glance, as these works provide theoretical guarantees for each group in terms of the number of datapoints from that group.
Multi-group learning \cite{rothblum_multi-group_2021, tosh_simple_2022} considers a similar problem to multi-distribution learning; by assuming that there exists a hypothesis that is simultaneously $\epsilon$-optimal on every distribution (an assumption not made in our setting) they compare their learned hypothesis against the best hypothesis for each individual distribution.

\paragraph{Multi-source domain adaptation.}
Multi-source domain adaptation, or multi-task learning, is another related line of work that is concerned with using data from multiple different training distributions to learn some target distribution, under the assumption that the training and target distributions share some task relatedness~\cite{ben-david_exploiting_2003,mansour_domain_2008}.
Multi-distribution learning can be framed similarly as using a finite set of training distributions to simultaneously learn the convex hull of the training distributions.
Interestingly, the requirement in the multi-distribution setting of learning the entire convex hull obviates the need for the task-relatedness assumptions of multi-source learning.

\paragraph{Stochastic game equilibria.}
Our approach relates to a line of research on using online algorithms to find min-max equilibria by playing no-regret algorithms against one another~\cite{robinson_iterative_1951,freund_decision-theoretic_1997,rakhlin_optimization_2013,daskalakis_near-optimal_2011,daskalakis_near-optimal_2021}.
Online mirror descent (OMD) is a well-studied family of methods that can find approximate minima of convex functions, and also find approximate min-max equilibria of convex-concave games, with high probability, using noisy first-order information~\cite{robbins_stochastic_1951, nemirovskij_problem_1983,hart_simple_2000,beck_mirror_2003}.
We bring these online learning tools to bear on the problem of finding saddle points in robust optimization formulations.
The primary technical difference between multi-distribution learning and traditional saddle-point optimization problems is that we have sample access to data distributions instead of noisy local gradients.

\paragraph{Other paradigms.}
Several other machine learning paradigms also consider learning from multiple distributions.
Notably, \emph{distributed learning} (e.g.,~\cite{rabbat_quantized_2005,boyd_distributed_2011,balcan_distributed_2012,daume_efficient_2012,shamir_communication-efficient_2014}) and \emph{federated learning} (e.g.,~\cite{konecny_federated_2016, konecny_federated_2016-1, mcmahan_communication-efficient_2017}) consider learning from data that is spread across multiple sources or devices.
Classically, both of these settings have focused on minimizing the training or testing error \emph{averaged} over these devices.
The literature in these fields has primarily focused on methods for minimizing the average loss using communication-efficient, private, and robust-to-dropout training methods.
However, optimizing average performance produces models that can significantly underperform on some data sources, especially when the data is heterogeneously spread across the sources.
In comparison, multi-distribution learning paradigms such as collaborative learning~\cite{blum_collaborative_2017}, agnostic federated learning~\cite{mohri_agnostic_2019}, and Group DRO~\cite{sagawa_distributionally_2020} learn models that perform well across any one of the data sources.

\paragraph{Subsequent work.}
\citet{haghtalabmulticalibration} formalized multicalibration as a type of multi-distribution learning, building on the framework presented in this manuscript.
Their work improves upon state-of-art multicalibration algorithms by implementing multi-distribution learning game dynamics using online learning algorithms that leverage the structure of calibration losses.
\citet{zhangStochasticApproximation2023} extended the discussion on the sample complexity of Group DRO to settings with data budgets.
They also noted an erroneous bandit-to-full-information reduction in an earlier version of this manuscript, which we corrected in a previous version (arXiv V2) with a minor change that employs Exp3 \cite{neu15} or ELP \cite{alonFromBanditsToExperts2013} in place of our earlier reduction.
\citet{haghtalabcolt} presented steps towards answering the sample complexity of multi-distribution learning with VC classes. 
This open problem was recently settled up to log factors by \citet{zhang2023optimal,peng2023sample}.

\section{Preliminaries}
\label{section:preliminaries}

Throughout this manuscript, we use the shorthands $\tsv{x}{1:T} \asseq \tsv{x}{1}, \dots, \tsv{x}{T}$ and $f(\cdot, b) \asseq a \mapsto f(a, b)$.
We write $\simplex(\cA)$ to denote the set of probability distributions supported on a set $A$ and $\simplex_d$ to denote the probability simplex in $\reals^{d-1}$.
We use $\norm{\cdot}_*$ to denote the dual of the norm $\norm{\cdot}$ and $e_i \in \reals^\numactions$ to denote the $i$th standard basis vector.
Given a data distribution $\dist$ supported on the space of datapoints $\datapoints$, hypothesis class $\hyps$, and a loss function $\loss: \hyps \times \datapoints \to [0, 1]$, we denote the expected loss (risk) of a hypothesis $\hyp \in \hyps$ by $\risk_{\dist, \loss}(\hyp) \asseq \EEs{z \sim \dist}{\loss(\hyp, z)}$, writing $\risk_\dist(\hyp)$ if $\loss$ is clear from context.

\subsection{Multi-Distribution Learning}
The goal of multi-distribution learning is finding a hypothesis that uniformly minimizes expected loss across multiple data distributions and loss functions.
Importantly, we make no assumptions on the relationships between the data distributions; for example, we do not assume the existence of a hypothesis that is simultaneously optimal for every distribution.
Formally, given a set of data distributions $\dists = \bsetf{\dist_i}_{i=1}^\numdists$, losses $\losses = \bsetf{\loss_j}_{j=1}^\numlosses$, and a hypothesis class $\hyps$, we say a hypothesis $\hyp$ is $\epsilon$-optimal for the multi-distribution learning problem $(\dists, \losses, \hyps)$ if
\begin{align}
	\label{eq:deterministic_multidist_learning_goal}
	\max_{\dist \in \dists}  \max_{\loss \in \losses} \risk_{\dist, \loss}({\hyp}) \leq \detopt + \epsilon,
	\text{ where } \detopt \asseq \min_{\hyp \in \hyps} \max_{\dist \in \dists} \max_{\loss \in \losses} \risk_{\dist, \loss}({\hyp}).
\end{align}
\noindent
Throughout this manuscript, we will often assume we are working with smooth and convex loss functions.
Formally, we say a multi-distribution learning problem $(\dists, \losses, \hyps)$ has smooth convex losses if two conditions are met.
First, $\hyps$ is parameterized by a convex compact Euclidean parameter space $\params$ such that $\hyps = \bset{\hyp_\theta}_{\theta \in \params}$.
Second, for the same parameter space $\params$, for every loss function $\loss \in \losses$ and datapoint $z \in \datapoints$, the mapping $f: \Theta \to [0, 1]$ defined as $f(\theta) = \loss(\hyp_\theta, z)$ is convex and $1$-smooth; i.e., $\norm{\nabla_{\theta} f(\theta)} \leq 1$ for all $\theta \in \Theta$.
We remark that the assumption of smooth convex losses is a weak assumption.
In fact, we will observe that our results on smooth convex losses easily extend to bounded non-smooth non-convex losses when the hypothesis class $\hyps$ is finite or combinatorially bounded, such as when $\hyps$ has finite VC dimension or Littlestone dimension \cite{littlestone_learning_1987}.

\paragraph{Sample complexity.}
We are interested in the design of multi-distribution learning algorithms that have sample access to the distributions $\dist_1, \dots, \dist_\numdists$ and only take a small number of samples from these distributions overall.
We formalize this access by defining a set of \emph{example oracles}, $\exoracle(\dist_1), \dots, \exoracle(\dist_\numdists)$, where each $\exoracle(\dist_i)$ returns i.i.d.\ samples from $\dist_i$.
We can then define the sample complexity of a multi-distribution learning algorithm by the cumulative number of calls it makes to these example oracles in order to find a solution. 

We note that a multi-distribution learning algorithm may make these example oracle calls in an adaptive fashion; i.e., choosing which example oracle to call based on the datapoints it received from previous oracle calls.
As first noted by \citet{blum_collaborative_2017}, this ability to query for samples \emph{on-demand} is critical for achieving efficient multi-distribution learning sample complexities.
We also note that multi-distribution algorithms can use a set of example oracles to sample from any mixture distribution $\maxweight \in \simplex{\dists}$; e.g., by first sampling a supporting distribution $\dist_i$ from the mixture distribution and then calling its example oracle $\exoracle(\dist_i)$.

\subsection{Instances of Multi-Distribution Learning}
Multi-distribution learning unifies the problem formulations of collaborative learning~\cite{blum_collaborative_2017}, agnostic federated learning~\cite{mohri_agnostic_2019}, and group distributionally robust optimization (group DRO)~\cite{sagawa_distributionally_2020}.
These problems have each spawned a line of highly influential works but were previously not recognized to be equivalent.
We emphasize our view that multi-distribution learning is a particularly useful level of generality at which to study these problems, as it allows for their unified treatment both conceptually and algorithmically.

\paragraph{Collaborative learning.}
In the \emph{collaborative PAC learning model} of~\citet{blum_collaborative_2017}, and its agnostic extensions by~\citet{nguyen_improved_2018}, the goal is to learn a hypothesis that guarantees small risk for every distribution in a collection of distributions.
These data distributions are usually interpreted as the heterogeneous problem domains faced by multiple participants that are collaborating on data collection; the goal of collaborative learning is to learn a machine learning model that all participants are satisfied with.

Collaborative learning is usually studied in a supervised learning setting where datapoints consist of a feature-label pair, i.e., $\datapoints = \features \times \labels$, and where hypothesis classes $\hyps \subset \labels^\features$ are either finite or combinatorially bounded.
Importantly, loss functions are assumed to be bounded in $[0, 1]$, but may be non-smooth and non-convex.
Formally, given a set of data distributions, $\dists \asseq \bsetf{ \dist_1, \dots, \dist_\numdists}$, supported on $\features \times \labels$, a loss function $\loss: \labels^\features \times \datapoints \to [0, 1]$, and a hypothesis class $\hyps \subset \labels^\features$, a collaborative learning instance, $(\hyps, \dists)$, is formulated as the problem of finding a solution $\hyp \in \labels^\features$ such that
\begin{align}
	\label{eq:deterministic_collab_learning_goal}
	\max_{\dist \in \dists} \risk_\dist({\hyp}) \leq \detopt + \epsilon,
	\text{ where } \detopt \asseq \min_{\hyp \in \hyps} \max_{\dist \in \dists} \risk_\dist({\hyp}).
\end{align}
We say a solution $\hyp$ is \emph{proper} if it is in class, i.e., $\hyp \in \hyps$, and \emph{randomized} if $\hyp$ is a probability distribution supported on the class, i.e., $\hyp \in \simplex(\hyps)$.
In the latter case, we define the expected loss for a randomized hypothesis as $\risk_\dist(\hyp) \asseq \EEsf{f \sim \hyp}{\risk_\dist(\hhyp)}$.

Multi-distribution learning with smooth convex losses and collaborative learning seem to differ significantly in terms of their formally definition.
However, we can reduce any collaborative learning problem to multi-distribution learning with smooth convex loss functions---as long as we allow for improper or randomized solutions to our collaborative learning problem.
Allowing for improper or randomized solutions is not unreasonable and is in fact necessary to achieve non-trivial sample complexities in collaborative learning \cite{blum_collaborative_2017}.

The first step to reducing collaborative learning to multi-distribution learning is to relax the optimization problem on the hypothesis class $\hyps$ onto the class of randomized hypotheses $\simplex(\hyps)$.
\begin{fact}
    \label{fact:collablearning}
    Consider a collaborative learning problem $(\hyps, \dists)$.
    Define the relaxed loss function $\tilde \loss: \simplex(\hyps) \to [0, 1]$ as $\tilde \loss (\hyp, z) = \EEs{f \sim \hyp}{\loss(f, z)}$.
    The induced losses in the multi-distribution learning problem, $(\dists, \bset{\tilde \loss}, \simplex(\hyps))$, are smooth and convex, and any $\epsilon$-optimal solution $\hyp \in \simplex(\hyps)$ is also an $\epsilon$-optimal randomized solution to the collaborative learning problem $(\hyps, \dists)$; i.e., it satisfies Equation~\ref{eq:deterministic_collab_learning_goal}.
\end{fact}
This fact implies that multi-distribution learning can solve for deterministic but improper solutions to collaborative learning problems.
This is because we can always extract a deterministic solution from a non-deterministic solution $\hyp \in \simplex(\hyps)$ by taking a majority vote, where we denote the majority vote hypothesis as ${\hyp}_\text{Maj}$.
The expected loss guarantee of this deterministic hypothesis is approximately bounded by that of the randomized ${\hyp}$.
We state this formally below for the setting where $\hyps$ is a set of binary classifiers; that is, where the label space is binary, $\labels = \bset{0, 1}$, and the loss function $\loss$ can be written as $\loss(\hyp, (x, y)) = g(\hyp(x), y)$ for some choice of $g: \labels^2 \to [0,1]$.

\begin{restatable}{fact}{randomizedtodet}
\label{fact:randomizedtodet}
Consider a collaborative learning problem $(\hyps, \dists)$ on a set of binary classifiers.
For any randomized solution $\hyp \in \simplex(\hyps)$, define the deterministic hypothesis $\hyp_\text{Maj}$ as $\hyp_\text{Maj}(x) = 1[\Pr_{f \sim \hyp}(f(x) = 1) > \frac 12]$.
The expected loss of $\hyp^{Maj}$ is bounded by $\max_{\dist \in \dists} \risk_\dist(\hyp_\text{Maj}) \leq 2 \max_{\dist \in \dists} \risk_\dist({\hyp})$.
\end{restatable}

\paragraph{Group distributionally robust optimization.}
In the closely related setting of \emph{group distributionally robust optimization (group DRO)} of~\citet{sagawa_distributionally_2020}, the goal is similarly to learn some hypothesis that guarantees small risk for every data distribution in a collection of distributions.
In group DRO, the various data distributions are usually interpreted to either represent heterogeneous user populations and protected groups (for algorithmic fairness applications) or potential domains in which a model may be deployed (for robustness applications). 

In contrast to the collaborative learning problem, group DRO problems are typically studied in a convex optimization setting where the hypothesis class is parameterized by some convex set and the loss function is smooth and convex.
That is, the usual definition of the group DRO problem setting coincides with the definition of multi-distribution learning with \emph{a single} smooth convex loss, i.e., $\setsize{\losses} = 1$.
Unlike in collaborative learning where we are interested in potentially improper or randomized solutions, the goal of group DRO is to learn a proper model $\hyp_\theta \in \hyps$ where
\begin{align}
	\label{eq:dro}
	\max_{\dist \in \dists} \risk_{\dist}(\hyp_\theta) \leq \opt + \epsilon,
	\text{ where } \opt \asseq \min_{\theta^* \in \Theta} \max_{\dist \in \dists}  \risk_{\dist}(\hyp_{\theta^*}).
\end{align}
It is the convexity of the group DRO problem setting that allows for the efficient learning of proper solutions and avoid relaxation to randomized solutions.

\paragraph{Agnostic federated learning.}
The agnostic federated learning framework of~\citet{mohri_agnostic_2019} also coincides with multi-distribution learning with a single loss function.
Like group distributionally robust optimization, agnostic federated learning is usually studied in a convex optimization setting with convex parameter spaces and smooth convex losses.
As the general agnostic federated learning setting does not differ from group distributionally robust optimization in its formulation, we provide an identical treatment of both settings in Section~\ref{section:gdroupperbound}.

\subsection{Technical Background}
We will use tools and definitions from the literature on zero-sum games and no-regret learning throughout the paper.
This section provides a brief overview of these concepts.

\paragraph{Zero-sum games.}
A two-player zero-sum game is described by the tuple $(\minactions, \maxactions, \payoff)$ where $\minactions, \maxactions$ are convex compact action sets and $\payoff: \minactions \times \maxactions \rightarrow [0, 1]$ is the game payoff.
The player who chooses from $\minactions$ is called the minimizing player and tries to minimize the game payoff $\payoff$, while the player who chooses from $\maxactions$ is called the maximizing player.
A pair of actions $(\mina, \maxa)$ is called an \emph{$\epsilon$-min-max} equilibrium if neither player can unilaterally improve their objective by more than $\epsilon$; that is, $\payoff(\mina, \maxa) - \min_{\mina^* \in \minactions} \payoff(\mina^*, \maxa) \leq \epsilon$ and $\max_{\maxa^* \in \maxactions} \payoff(\mina, {\maxa^*}) - \payoff(\mina, \maxa) \leq \epsilon$.
If $\phi$ is convex-concave---i.e., $\phi(\cdot, \maxa)$ is convex for every $\maxa \in \maxactions$ and $\phi(\mina, \cdot)$ is concave for every $\mina \in \minactions$---then an $\epsilon$-min-max equilibrium always exists for every $\epsilon \geq 0$.
In the next section, we will describe methods that find $\epsilon$-min-max equilibria by playing online learning algorithms against each other, a technique known as \emph{no-regret game dynamics}~\cite{freund_decision-theoretic_1997}.

\paragraph{No-regret learning.}
A no-regret (or \emph{online}) learning algorithm $\learningalgorithm_\actions$ maps from a sequence of costs $\tsv{\cost}{1:t-1}$ to an action $\tsv{\action}{t} \in \actions$, where $\tsv{\action}{t} = \learningalgorithm_\actions(\tsv{\cost}{1:t-1})$.
Notationally, we use the subscript $\actions$ when writing an online learning algorithm $\cQ_\actions$ to denote the action set that the algorithm $\cQ_\actions$ is defined to act on.
Regret is defined for a sequence of actions $\tsv{\action}{1}, \dots, \tsv{\action}{T} \in \actions$ and costs $\tsv{\cost}{1}, \dots, \tsv{\cost}{T} : \actions \to [0, 1]$ as follows: \[\regret(\tsv{\action}{1:T}, \tsv{\cost}{1:T}) \asseq \sum_{t=1}^T \tsv{\cost}{t}(\tsv{\action}{t}) - \min_{\action^* \in \actions} \sum_{t=1}^T \tsv{\cost}{t}(\action^*).\]
We say that a no-regret learning algorithm $\learningalgorithm_\actions$ has a regret guarantee of $\gamma_T(\learningalgorithm_\actions)$ if, for any sequence of linear cost functions $\tsv{\cost}{1:T}$ of bounded norm, i.e., $\max_{t \in [T]} \norm{\tsv{\cost}{t}} \leq 1$, the algorithm $\learningalgorithm_\actions$ chooses an action sequence $\tsv{\action}{1:T}$ with the regret bound $\regret(\tsv{\action}{1:T}, \tsv{\cost}{1:T}) \leq \sqrt{\gamma_T(\learningalgorithm_\actions) T}$.

\paragraph{Examples of no-regret algorithms on probability simplexes.}
A well-studied online learning setting is that in which the action set is a probability simplex,$\actions = \simplex_n$, and all costs are linear functions of bounded norm.
In this setting, we can interpret online learning algorithms as choosing mixed strategies $\tsv{\action}{t} \in \simplex_n$ over a set of \emph{meta-actions}, $\bset{1, \dots, n}$, and the adversary as assigning a cost $\bset{\tsv{c}{t}(e_1), \dots, \tsv{c}{t}(e_n)}$ to each meta-action, so that the algorithm incurs the cost $\EEs{i \sim \tsv{\action}{t}}{\tsv{c}{t}(e_i)}$.
An example of a no-regret algorithm in this setting is Exponential Gradient Descent (Hedge), defined as
\begin{align}
	\label{eq:proxmapping}
	\hedgealgorithm_{\actions}(\tsv{\cost}{1:t-1}) \asseq \tsv{\tilde{\action}}{t} /\norm{\tsv{\tilde{\action}}{t}}_1
	\text{ where } \tilde{\action} \in \reals^\numactions \text{ and }
	\tilde{\action}_i
	\asseq
	\exp \para{-\stepsize \sum_{\tau=1}^{t-1} \tsv{\cost}{\tau}(e_i)},
\end{align}
where $\stepsize \in (0, 1)$ is a learning rate.
The following lemma states a classical result for exponential gradient descent (Hedge), showing a regret guarantee of $\bigO{\log(\numactions)}$.
\begin{restatable}[\cite{vishnoi_algorithms_2021}]{lemma}{egdconvergence}
	\label{lemma:egdconvergence}
	Let $\tsv{\cost}{1:T}$ be any linear cost sequence where $\max_{t \in [T]} \norm{\tsv{\cost}{t}}_\infty \leq 1$ and $\actions = \simplex_\numactions$.
	When $\stepsize = \sqrt{ \log (\numactions / T)}$, the actions $\tsv{\action}{1:T}$ chosen by Hedge satisfy $\smash{\regret(\tsv{\action}{1:T}, \tsv{\cost}{1:T})
			\leq 2 \sqrt{\log(\numactions) / T}}$.
\end{restatable}
\noindent
There also exist \emph{partial feedback} no-regret algorithms---also known as semi-bandit algorithms---that only need to observe the cost functions at each timestep for a few meta-actions (i.e., along a few basis vectors).
We can formalize these partial feedback (semi-bandit) algorithms as returning not only an action $\tsv{\action}{t} \in \simplex_\numactions$ at each timestep $t$ but also returning the meta-actions $\tsv{I}{t} \subseteq [\numactions]$ whose costs it will observe.
We can therefore, somewhat unconventionally, write these algorithms as a mapping
\begin{align*}
\bsetf{\tsv{\cost}{1}(e_i)}_{i \in \tsv{I}{1}}, \dots, \bsetf{\tsv{\cost}{t-1}(e_i)}_{i \in \tsv{I}{t-1}} \mapsto \tsv{\action}{t}, \tsv{I}{t}.
\end{align*}
The well-known partial feedback algorithm Exp3 chooses $\tsv{\action}{t} = \hedgealgorithm(\tsv{\tilde{\cost}}{1:t-1})$ and $\tsv{I}{t} = \bsetf{\tsv{i}{t}}$ at each timestep, where $\tsv{i}{t} \simiid \tsv{\action}{t}$ and $\tsv{\tilde{\cost}}{t}(\action) = \action_{\tsv{i}{t}} \tsv{\cost}{t}(e_{\tsv{i}{t}}) / (\tsv{\action}{t}_{\tsv{i}{t}} + \lambda)$ and where $\lambda \geq 0$ is a stepsize \cite{neu15}.
An alternatie partial feedback algorithm is ELP which, when given a partition $P$ of the meta-actions $[n]$ into $k$ subsets, guarantees $\tsv{I}{t} \in P$ at each timestep.
That is, it fixes a grouping of the meta-actions a priori and at each timestep only observes the costs of meta-actions belonging to a particular group.
\begin{lemma}[\cite{mannor2011bandits}]
	\label{lemma:elpconvergence}
	Let $\tsv{\cost}{1:T}$ be arbitrary linear costs where $\max_{t \in [T]} \norm{\tsv{\cost}{t}}_\infty \leq 1$ and $\actions = \simplex_\numactions$.
	For any $\delta \in (0, 1)$ and partition $P$ of $[n]$, the actions $\tsv{\action}{1:T}$ chosen by ELP satisfy $\smash{\regret(\tsv{\action}{1:T}, \tsv{\cost}{1:T})
			\leq 2 \sqrt{ \setsize{P} \log(\numactions / \delta) / T}}$ with probability $1 - \delta$.
   Moreover, only cost components from one element of $P$ are observed per timestep: $\setsize{\tsv{I}{t}} \in P$.
\end{lemma}
We emphasize that the results in this manuscript are stated to accommodate general choices of online learning algorithms, with different guarantees and tradeoffs arising depending on which specific online learning algorithms one employs.

\section{Overview of Our Approach}
\label{sec:tech-overview}

In this section, we provide an overview of our general approach for studying the sample complexity of multi-distribution learning.
Our approach consists of two steps: (1) reducing multi-distribution learning to the problem of finding the equilibrium of a convex-concave zero-sum game, and (2) implementing game dynamics to efficiently find an equilibrium using only stochastic feedback.

\subsection{Multi-Distribution Learning as a Zero-Sum Game}
The multi-distribution learning problem corresponds to a zero-sum game with a minimizing player having action set $\hyps$, a maximizing player having action set $\dists \times \losses$, and a payoff function $\phi(\hyp, (\dist, \loss)) = \risk_{\dist, \loss}(\hyp)$.
Intuitively, the minimizing player can be interpreted as a learner who proposes candidate solutions while the maximizing player can be interpreted as an auditor who tries to pick a data distribution and loss function for which the learner's hypothesis performs poorly.
It is not hard to see that any $\epsilon$-min-max equilibrium $(\hyp, \dist)$ of this game corresponds to a $2\epsilon$-optimal solution.
\begin{fact}
	\label{fact:equilsolution}
    Given a multi-distribution learning problem, $(\dists, \losses, \hyps)$, define the zero-sum game $(\minactions, \maxactions, \payoff)$ where $\minactions = \hyps$, $\maxactions = \dists \times \losses$, and $\payoff(\mina, \maxa) = \risk_{\maxa}(\mina)$.
	In any $\epsilon$-min-max equilibrium $(\mina, \maxa)$, $\mina$ is a $2 \epsilon$-optimal solution.
\end{fact}
\begin{proof}
	If $({\mina}, {\maxa})$ is an $\epsilon$-min-max equilibria, the following holds by definition
	\begin{align*}
		\risk_{{\maxa}}({\mina}) & \leq \min_{\hyp^* \in \hyps} \risk_{{\maxa}}(\hyp^*) + \epsilon
		\text{ and } \risk_{{\maxa}}({\mina}) \geq \max_{\dist^* \in \dists, \loss^* \in \losses} \risk_{\dist^*, \loss^*}({\mina})- \epsilon.
	\end{align*}
	Rearranging gives $\max_{\dist^* \in \dists, \loss^* \in \losses} \risk_{\dist^*, \loss^*}({\mina}) \leq \min_{\hyp^* \in \hyps} \risk_{{\maxa}}(\hyp^*) + 2\epsilon \leq \opt + 2\epsilon$.
\end{proof}
\noindent
A multi-distribution learning problem $(\dists, \losses, \hyps)$ with \emph{convex losses} can similarly be written as a convex-concave zero-sum game where a minimizing player chooses from the actions $\Theta$, a maximizing player chooses from the actions $\dists \times \losses$, and the payoff function is defined as $\phi(p, q) = \risk_q(h_p)$.
As we previously noted, as the payoff function is convex-concave, a min-max equilibrium of this game must exist.

Many tools have been developed for efficiently finding the min-max equilibria of convex-concave zero-sum games.
The connection between multi-distribution learning and zero-sum games allows us to draw on these tools to derive efficient learning algorithms.

\paragraph{Unknown payoff functions.}
The main challenge we will encounter is that of efficiently estimating the payoff function of the multi-distribution learning game, given that evaluating the function $\phi(p, q) = \risk_q(h_p)$ requires computing expectations for an unknown data distribution.
Typically, to compute the min-max equilibrium of a convex-concave game, one needs a first-order approximation for the payoff function $\phi$ at various strategy profiles---that is, we require the gradients $\nabla_p \phi(p, q)$ and $\nabla_q \phi(p, q)$ for various choices of actions $p \in \minactions$ and $q \in \maxactions$.
We will achieve this by designing \emph{noisy first-order oracles} that, when queried with a strategy profile $(p, q)$, return unbiased estimates of the gradient $\nabla_p \phi(p, q)$ or $\nabla_q \phi(p, q)$.
To control the variance of these oracles, we will also ask that their estimates be bounded in norm, as we will formalize in the sequel.
Behind the scenes, we will implement these first-order approximations by querying example oracles.

A complication to implementing these noisy first-order oracles efficiently is that payoff estimation is more costly for the maximizing player than the minimizing player. Indeed, consider a strategy profile $(p, q)$ in the multi-distribution learning game.
Obtaining an unbiased bounded estimate of the minimizing player (learner)'s payoff gradient requires only drawing a single datapoint from the mixture distribution specified by the other player (the auditor), since the learner only needs a counterfactual estimate of how well each hypothesis would have performed on the mixture.
However, obtaining an unbiased bounded estimate of the maximizing player (the auditor)'s payoff gradient requires drawing $\numdists$ datapoints, since the auditor needs a counterfactual estimate of how well the minimizing player's hypothesis would have performed on each potential data distribution $\dist_1, \dots, \dist_\numdists$.
This intuitive arugment is formalized as follows.
\begin{fact}
	\label{fact:estimators}
	Consider a multi-distribution learning problem $(\dists, \losses, \hyps)$ with 1-smooth losses and a strategy profile $\hyp_\theta \in \hyps$ and $q \in \simplex(\dists \times \losses)$.
	The gradient vector $\nabla_\theta {{\loss}(\hyp_\theta, z)}$, where $z \;\simiid\; {\dist}$ and $(\dist, \loss) \;\simiid\; q$, is an unbiased bounded estimate of the first-order information $\nabla_\theta \risk_{q}(\hyp_\theta)$; i.e., $\EEs{z \sim \dist, (\dist, \loss) \sim q}{\nabla_\theta {{\loss}(\hyp_\theta, z)}} = \nabla_\theta \risk_{q}(\hyp_\theta)$ and $\norm{\nabla_\theta {{\loss}(\hyp_\theta, z)}} \leq 1$.
	Similarly, the vector $[1 - \loss_j({\hyp_\theta}, z_i)]_{i \in [\numdists], j \in [\numlosses]}$ where $z_i \;\simiid\; \dist_i$ is an unbiased bounded estimate of the first-order information vector $1 - \nabla_q \risk_{q}(\hyp_\theta)$.
\end{fact}

\subsection{Equilibrium Computation in Stochastic Convex-Concave Games}
\begin{algorithm}[h]
\caption{Finding Equilibria in Convex-Concave Games with Asymmetric Costs.}
\label{alg:finite_general}
\begin{algorithmic}
\STATE \textbf{Input:} Action sets $\minactions, \maxactions$, steps $T$, first-order oracles $\minop, \maxop$, and online learning algorithms $\learningalgorithm_{\minactions},\learningalgorithm_{\maxactions}$;
\FOR{$t = 1, 2, \dots, T$}
\STATE Let $\tsv{\mina}{t} = \learningalgorithm_{\minactions}\para{\bset{\mina \mapsto \inner{\minop(\tsv{\mina}{\tau}, \tsv{\maxa}{\tau})}{\mina}}_{\tau \in [t-1]}}$;
\STATE Let $\tsv{\maxa}{t} = \learningalgorithm_{\maxactions}\para{\bset{\maxa \mapsto \inner{\maxop(\tsv{\mina}{\tau}, \tsv{\maxa}{\tau})}{\maxa}}_{\tau \in [t-1]}}$;
\ENDFOR
\STATE Return ${\overline{\mina}} = \frac{1}{T} \sum_{t=1}^T \tsv{\mina}{t}$ and ${\overline{\maxa}} = \frac{1}{T} \sum_{t=1}^T \tsv{\maxa}{t}$;
\end{algorithmic}
\end{algorithm}
We now describe an online learning framework for finding equilibria in stochastic games using game dynamics.
We will later see this framework, which is described by Algorithm~\ref{alg:finite_general}, easily accommodates the asymmetric costs of estimating each player's payoff gradients.
Lemma~\ref{lemma:zerosumasymmetricfull} outlines a guarantee of returning an approximate min-max equilibrium with high probability.
We note that the guarantee of Lemma~\ref{lemma:zerosumasymmetricfull} is stated in terms of the regret bounds $\gamma_T(\cQ_{\minactions})$ and $\gamma_T(\cQ_{\maxactions})$ of the online learning algorithms that we plug into Algorithm~\ref{alg:finite_general}.
This means that choosing different online learning algorithms to be $\cQ_{\minactions}$ and $\cQ_{\maxactions}$ will yield different guarantees.
\begin{restatable}{lemma}{zerosumasymmetricfull}
    \label{lemma:zerosumasymmetricfull}
    Consider a convex-concave zero-sum game  $(\minactions, \maxactions, \payoff)$ with an $\lipschitz$-smooth payoff $\phi$.
    Assume that
    \begin{enumerate}
        \item $\minop$ is a noisy first-order oracle that returns unbiased, bounded, and independent estimates of $\nabla_\mina \phi(\mina, \maxa)$. That is, for all $\mina \in \minactions$ and $\maxa \in \maxactions$, we have $\norm{\minop(\mina, \maxa)} \leq \lipschitz$  with probability one and $\EE{\minop(\mina, \maxa)} = \nabla_\mina \phi(\mina, \maxa)$.
        \item $\maxop$ is a noisy first-order oracle that returns unbiased, bounded, and independent estimates of $-\nabla_\maxa \phi(\mina, \maxa)$.
        \item The action sets $\minactions$ and $\maxactions$ have a diameters of at most $R$ in the dual norm $\norm{\cdot}_*$, i.e. $\max_{p, p' \in \minactions} \norm{p - p'}_* \leq {R}$ and $\max_{q, q' \in \maxactions} \norm{q - q'}_* \leq {R}$.
    \end{enumerate}
    Then Algorithm~\ref{alg:finite_general} returns an $\epsilon$-min-max equilibrium with probability $1-\delta$ if
    \begin{align}
        \label{eq:lemmasample}
        T \geq &  \frac{4 \lipschitz^2}{\epsilon^2}
        \para{32 R^2 \log(2/ \delta) + 25 \gamma_T (\learningalgorithm_{\minactions}) + 25 \gamma_T (\learningalgorithm_{\maxactions})
        }.
    \end{align}
\end{restatable}
\noindent
Informally, we can interpret the regret bound $\gamma_T (\learningalgorithm_{\minactions})$ as the difficulty of making rational choices for the minimizing player and $\gamma_T (\learningalgorithm_{\maxactions})$ as the difficulty of making rational choices for the maximizing player.
This lemma then says that the number of iterations required to find an equilibrium depends on the \emph{additive combination} of the complexity seen by each player.

Before we proceed to a proof of this lemma, we recall some standard results on game dynamics and online learning.
First, we recall that no-regret dynamics efficiently learns equilibria in convex-concave games~\cite{freund_decision-theoretic_1997}.
This fact implies that, in order to learn the equilibria of our multi-distribution learning game, it suffices to design suitable no-regret algorithms.
\begin{fact}
\label{fact:vierrtoequilibria}
Let $(\minactions, \maxactions, \payoff)$ be a convex-concave zero-sum game.
For any actions $\tsv{\mina}{1:T} \in \minactions$ and $\tsv{\maxa}{1:T} \in \maxactions$ with regret $\regret\para{\tsv{\mina}{1:T}, \bset{\payoff(\cdot, \tsv{\maxa}{t})}_{t \in [T]}} \leq T \epsilon$ and $\regret\para{\tsv{\maxa}{1:T}, \bset{  - \payoff(\tsv{\mina}{t}, \cdot)}_{t \in [T]}} \leq T \epsilon$, the average actions $\frac{1}{T} \sum_{t=1}^T \tsv{\mina}{t}$ and $ \frac{1}{T} \sum_{t=1}^T \tsv{\maxa}{t}$ form a $2\epsilon$-min-max equilibrium.
\end{fact}
\begin{proof}
    By convexity, $\overline{\mina}  \asseq \frac{1}{T} \sum_{t=1}^T \tsv{\mina}{t} \in \minactions$ and $\overline{\maxa}\asseq \frac{1}{T} \sum_{t=1}^T \tsv{\maxa}{t} \in \maxactions$.
    Since $\phi$ is concave in its second argument, we can apply Jensen's inequality to the regret bound of the minimizing player to get
    \begin{align*}
         \regret\para{\bsetf{\overline{\mina}}, \bset{\payoff(\cdot, \overline{\maxa})}}
         & = \phi(\overline{\mina}, \overline{\maxa}) - \min_{\mina^* \in \minactions} \phi(\mina^*, \overline{\maxa}) 
          \leq \frac 1T \sum_{t=1}^T \phi(\overline{\mina}, \tsv{\maxa}{t}) - \min_{\mina^* \in \minactions} \phi(\mina^*, \overline{\maxa}) \leq \epsilon.
    \end{align*}
    Since $\phi$ is convex in its first argument, we can again apply Jensen's inequality, this time to the regret bound of the maximizing player, to get
    \begin{align*}
         \regret\para{\bsetf{\overline{\maxa}}, \bset{-\payoff(\overline{\mina}, \cdot)}}
         & = \max_{\maxa^* \in \maxactions} \phi(\overline{\mina}, \maxa^*) - \phi(\overline{\mina}, \overline{\maxa})
          \leq \max_{\maxa^* \in \maxactions} \phi(\overline{\mina}, \maxa^*) - \frac 1T \sum_{t=1}^T \phi(\tsv{\mina}{t}, \overline{\maxa})
          \leq \epsilon.
    \end{align*}
    Summing these inequalities yields that $\max_{\maxa^* \in \maxactions} \phi(\overline{\mina}, \maxa^*) - \min_{\mina^* \in \minactions} \phi(\mina^*, \overline{\maxa})  \leq 2 \epsilon$.
\end{proof}

Next, we recall that, in a no-regret learning problem with linear costs $\tsv{\cost}{1:T}$, a player can run any online learning algorithm directly on independent, unbiased, bounded estimates $\tsv{\hat{\cost}}{1:T}$ of its costs $\tsv{\cost}{1:T}$ and expect only a constant factor increase in its worst-case regret bound.
This fact, which is classical in both optimization theory \cite{nemirovskij_problem_1983, juditsky_solving_2011} and online learning theory \cite{freund_decision-theoretic_1997}, follows by a standard martingale argument.
That no-regret learning algorithms generalize well on stochastic costs will mean that we can efficiently implement no-regret dynamics on stochastic games using noisy payoff observations that need only be unbiased and bounded.
Importantly, this means we do not need to obtain $\epsilon$-accurate estimates of each players' payoff at each iteration, which would make no-regret dynamics prohibitively expensive in terms of sample complexity.

In the sequel, given a linear cost function $\cost: \actions \to \reals$, we will abuse notation and use $\cost$ to also denote the vector such that $\cost(\action) = \inner{\cost}{\action}$ for all $\action \in \actions$.
We will also use $\norm{\cost}$ to denote the norm of the vector $\cost$.
\begin{restatable}{fact}{concentration}
\label{fact:concentrationnoise}
Let $\tsv{\hat\cost}{1:T}$ be independent, unbiased estimates of a set of linear costs $\tsv{\cost}{1:T}$, where $\norm{\tsv{\cost}{t}} \leq L$ and $\norm{\tsv{\hat\cost}{t}} \leq L$ at all steps $t \in [T]$.
Assume an action diameter of $R$, i.e.  $\max_{a, a' \in \actions} \norm{a - a'}_* \leq R$.
The actions $\tsv{\action}{t} = \learningalgorithm_\actions(\tsv{\hat\cost}{1:t-1})$ chosen by applying an online learning algorithm $\learningalgorithm_\actions$ to the estimated costs $\tsv{\hat\cost}{1:T}$ satisfies the following generalization bound with probability $1 - \delta$:
\begin{align}
    \label{eq:localvierr}
    \regret(\tsv{\action}{1:T}, \tsv{\cost}{1:T}) -  \regret(\tsv{\action}{1:T}, \tsv{\hat\cost}{1:T})
    & \leq 4 \lipschitz \sqrt{T} \para{R \sqrt{2 \log(1/\delta)} + \sqrt{\gamma_T(\learningalgorithm_\actions)}}.
\end{align}
\end{restatable}
\begin{proof}
    We first upper bound the generalization error by the regret of actions $\tsv{\action}{1:T}$ on the cost differences $\bset{\tsv{\cost}{t} - \tsv{\tilde \cost}{t}}$. Since $\max_{\action^* \in \actions} \sum_{t=1}^T \inner{\tsv{\cost}{t}}{\action^*} - \max_{\action^* \in \actions} \sum_{t=1}^T  \inner{\tsv{\hat\cost}{t}}{\action^*} \leq \max_{\action^* \in \actions} \sum_{t=1}^T  \inner{\tsv{\cost}{t} - \tsv{\hat\cost}{t}}{\action^*}$, we can bound generalization error $\Delta$ by
    \begin{align*}
        \Delta \asseq \regret(\tsv{\action}{1:T}, \tsv{\cost}{1:T}) - \regret(\tsv{\action}{1:T}, \tsv{\hat\cost}{1:T})
        \leq \regret(\tsv{\action}{1:T}, \tsv{\cost}{1:T} - \tsv{\hat\cost}{1:T}).
    \end{align*}
    The remainder of the proof is dedicated to bounding this regret term, which we can write explicitly as
    \[
    \regret(\tsv{\action}{1:T}, \tsv{\cost}{1:T} - \tsv{\hat\cost}{1:T}) = \max_{\action^* \in \actions} \sum_{t=1}^T \inner{\tsv{\action}{t} - \action^*}{\tsv{\cost}{t} - \tsv{\hat\cost}{t}}.
    \]
    We will ultimately control this regret term with a martingale argument, appealing to the fact that at each timestep the noisy costs we observe are unbiased even conditioned on previous cost observations.
    However, we first need to control the variational term $\action^*$, which we will do with a standard approach of introducing a shadow term $\tsv{\epsilon}{t} = \learningalgorithm_\actions(\bsetf{\tsv{\cost}{\tau} - \tsv{\hat\cost}{\tau}}_{\tau \in [t-1]})$.
    That is, $\tsv{\epsilon}{1:T}$ is the result of (hypothetically) running the online learning algorithm $\learningalgorithm_\actions$ on the cost sequences $\bsetf{\tsv{\cost}{\tau} - \tsv{\hat\cost}{\tau}}$.
    Adding and subtracting the shadow terms $\tsv{\epsilon}{1:T}$ from the inner product,
    \begin{align*}
        \max_{\action^* \in \actions} \sum_{t=1}^T \inner{\tsv{\action}{t} - \action^*}{\tsv{\cost}{t} - \tsv{\hat\cost}{t}}
        &= \max_{\action^* \in \actions} \sum_{t=1}^T \inner{\tsv{\action}{t} - \tsv{\epsilon}{t} + \tsv{\epsilon}{t} - \action^*}{\tsv{\cost}{t} - \tsv{\hat\cost}{t}} \\
&        = \sum_{t=1}^T \inner{\tsv{\action}{t} - \tsv{\epsilon}{t}}{\tsv{\cost}{t} - \tsv{\hat\cost}{t}} + \max_{\action^* \in \actions} \sum_{t=1}^T \inner{ \tsv{\epsilon}{t} - \action^*}{\tsv{\cost}{t} - \tsv{\hat\cost}{t}}.
    \end{align*}
    Since we constructed $\tsv{\epsilon}{1:T}$  with our online learning algorithm $\learningalgorithm_\actions$,  we obtain a regret guarantee for the action sequence $\tsv{\epsilon}{1:T}$ on the cost sequence $\bset{\tsv{\cost}{t} - \tsv{\hat\cost}{t}}$, which yields
    \begin{align}
        \label{eq:byconstructionbound}
        \max_{\action^* \in \actions} \sum_{t=1}^T \inner{\tsv{\epsilon}{t} - \action^*}{\tsv{\cost}{t} - \tsv{\hat\cost}{t}} \in 4 \lipschitz \sqrt{\gamma_T(\learningalgorithm_\actions) T}.
    \end{align}
    The term $4\lipschitz$ appears in this bound since $\tsv{\cost}{t}(\action) - \tsv{\hat\cost}{t}(\action) \in [-2\lipschitz, 2\lipschitz]$ must be normalized to $[0,1]$.

    It remains to bound $\sum_{t=1}^T \inner{\tsv{\action}{t} - \tsv{\epsilon}{t}}{\tsv{\cost}{t} - \tsv{\hat\cost}{t}}$. This expression is a martingale
   because, for each summand $\inner{\tsv{\action}{t} - \tsv{\epsilon}{t}}{\tsv{\cost}{t} - \tsv{\hat\cost}{t}}$, the left-hand side $\tsv{\action}{t} - \tsv{\epsilon}{t}$ is conditionally (on previous summands) independent of $\tsv{\cost}{t} - \tsv{\hat\cost}{t}$. 
    Formally, we define the filtration $\bsetf{\tsv{\cF}{t}}_{t=0}^T$ as the sigma algebra generated by $\bsetf{\tsv{\hat\cost}{t}}_{t=1}^T$.
    By construction, we know that $(\tsv{\action}{t} - \tsv{\epsilon}{t})$ is $\tsv{\cF}{t-1}$-measurable and thus $\inner{\tsv{\action}{t} - \tsv{\epsilon}{t}}{\tsv{\cost}{t} - \tsv{\hat\cost}{t}}$ is $\tsv{\cF}{t}$-measurable.
    Since $\tsv{\hat\cost}{t}$ is unbiased, we also have that $\EEc{\inner{\tsv{\action}{t} - \tsv{\epsilon}{t}}{\tsv{\cost}{t} - \tsv{\hat\cost}{t}}}{\tsv{\cF}{t-1}} = 0$.
    Finally, we can observe that the difference sequence of our martingale can be bounded with Holder's inequality as
    \[
    \abs{\inner{\tsv{\action}{t} - \tsv{\epsilon}{t}}{\tsv{\cost}{t} - \tsv{\hat\cost}{t}}} 
    \leq \norm{\tsv{\action}{t} - \tsv{\epsilon}{t}}_* \norm{\tsv{\cost}{t} - \tsv{\hat\cost}{t}}
    \leq 4 R \lipschitz.
    \]
	By the Azuma-Hoeffding inequality, we can thus bound, for any $\epsilon > 0$,
	\begin{align*}
		  \Pr\para{\sum_{t=1}^T \inner{\tsv{\action}{t} - \tsv{\epsilon}{t}}{\tsv{\cost}{t} - \tsv{\hat\cost}{t}}
		\geq \epsilon} 
		\leq \exp \para{-\frac{\epsilon^2}{32 T  R^2 \lipschitz^2}}.
	\end{align*}
    We can rewrite this as saying, with probability $1 - \delta$, that $\sum_{t=1}^T \inner{\tsv{\action}{t} - \tsv{\epsilon}{t}}{\tsv{\cost}{t} - \tsv{\hat\cost}{t}} \leq 4 R \lipschitz \sqrt{2T \log(1/\delta)}$.
    In combination with \equationref{\ref{eq:byconstructionbound}}, this inequality yields the desired bound on $\Delta$.
\end{proof}

We intend to apply online learning algorithms to our convex-concave games, where the payoff function is not necessarily linear.
To overcome the fact that the concentration result in Fact~\ref{fact:concentrationnoise} is specific to linear costs, we now turn to showing that one can linearize any online learning problem with convex costs.
That is, we can reduce online learning on differentiable convex costs to online learning on linear costs, allowing us to apply Fact~\ref{fact:concentrationnoise}.
Specifically, we will use the concept of variational error, which is usually defined as $\vierr(\tsv{\action}{1:T}, \tsv{\cost}{1:T}) \asseq \regret(\tsv{\action}{1:T}, \tsv{\tilde{\cost}}{1:T})$ where $\tsv{\tilde{\cost}}{t}(\action) = \inner{\action}{\nabla \tsv{\cost}{t}(\tsv{\action}{t})}$.
We now formalize the fact that variational error yields an upper bound on regret: $\vierr(\tsv{\action}{1:T}, \tsv{\cost}{1:T}) \geq \regret(\tsv{\action}{1:T}, \tsv{\cost}{1:T})$.
\begin{fact}
    \label{fact:saddlepointvi}
    Let $\tsv{\cost}{1:T}: \actions \rightarrow \reals$ be convex functions on a convex compact domain $\actions$ and let $\partial \tsv{\cost}{t}(\tsv{\action}{t})$ be a partial subgradient of $\tsv{\cost}{t}$ at $\tsv{\action}{t}$.
    For any sequence $\tsv{\action}{1:T} \in \actions$,
    \begin{align*}
        \regret(\tsv{\action}{1:T}, \tsv{\cost}{1:T})
        \asseq
        \sum_{t=1}^T \tsv{\cost}{t}(\tsv{\action}{t})
        - \min_{\action^* \in \actions} \sum_{t=1}^T \tsv{\cost}{t}(\action^*)
        \leq
        \vierr(\tsv{\action}{1:T}, \tsv{\cost}{1:T})
        \asseq
       \max_{\action^* \in \actions} \sum_{t=1}^T \innerproduct{ \partial \tsv{\cost}{t}(\tsv{\action}{t})}{\tsv{\action}{t} - \action^*}.
    \end{align*}
\end{fact}
\begin{proof}
    By the convexity of $\payoff$, $\sum_{t=1}^T \innerproduct{ \partial \tsv{\cost}{t}(\tsv{\action}{t})}{\tsv{\action}{t} - \action^*}
        \geq
        \sum_{t=1}^T \tsv{\cost}{t}(\tsv{\action}{t})
        -  \tsv{\cost}{t}(\action^*)$ for any fixed $\action^* \in \actions$.
\end{proof}

We now turn to proving Lemma~\ref{lemma:zerosumasymmetricfull}.
\begin{proof}[Proof of Lemma \ref{lemma:zerosumasymmetricfull}]
   Let $\tsv{\minx{\hat{c}}}{\tau}(\mina) = \inner{\minop(\tsv{\mina}{\tau}, \tsv{\maxa}{\tau})}{\mina}$ and $\tsv{\maxx{\hat{c}}}{\tau}(\maxa) = \inner{\maxop(\tsv{\mina}{\tau}, \tsv{\maxa}{\tau})}{\maxa}$ denote the cost functions that the online learning algorithms $\learningalgorithm_{\minactions}$ and $\learningalgorithm_{\maxactions}$ are given in Algorithm~\ref{alg:finite_general}.
   We first recall that the bounds on the regret for $\learningalgorithm_{\minactions}$ and $\learningalgorithm_{\maxactions}$ yield the empirical regret bounds
    \begin{align*}
        \regret\para{\tsv{\mina}{1:T}, \tsv{\minx{\hat{c}}}{1:T}}
        \leq \lipschitz \sqrt{\gamma_T(\learningalgorithm_{\minactions}) T}, \quad  \regret\para{\tsv{\maxa}{1:T}, \tsv{\maxx{\hat{c}}}{1:T}}
        \leq \lipschitz \sqrt{\gamma_T(\learningalgorithm_{\maxactions}) T},
    \end{align*}
    as $\tsv{\minx{\hat\cost}}{1:T}$ and  $\tsv{\maxx{\hat\cost}}{1:T}$ are linear cost functions with a bounded norm of at most $L$.
    By Fact~\ref{fact:concentrationnoise}, with probability $1 - 2\delta$, we can bound the generalization error with the true cost functions $\tsv{\minx{c}}{\tau}(\mina) = \inner{{\nabla}_{\tsv{\mina}{\tau}} \payoff(\tsv{\mina}{\tau}, \tsv{\maxa}{\tau})}{\mina}$ and $\tsv{\maxx{c}}{\tau}(\maxa) = -\inner{{\nabla}_{\tsv{\maxa}{\tau}} \payoff(\tsv{\mina}{\tau}, \tsv{\maxa}{\tau})}{\maxa}$ by
    \begin{align*}
        \regret\para{\tsv{\mina}{1:T}, \tsv{\minx{\hat{c}}}{1:T}} - 
        \regret\para{\tsv{\mina}{1:T}, \tsv{\minx{{c}}}{1:T}} \leq 4  \lipschitz \sqrt{T} \para{R \sqrt{2 \log(1/\delta)} + \sqrt{\gamma_T(\learningalgorithm_{\minactions})}},  \\
        \regret\para{\tsv{\maxa}{1:T}, \tsv{\maxx{\hat{c}}}{1:T}} - 
        \regret\para{\tsv{\maxa}{1:T}, \tsv{\maxx{{c}}}{1:T}}   \leq 4 \lipschitz \sqrt{T} \para{R \sqrt{2 \log(1/\delta)} + \sqrt{\gamma_T(\learningalgorithm_{\maxactions})}}.
    \end{align*}
    Next, we observe that the costs $\minx{\cost}$ and $\maxx{\cost}$ are constructed so that regret on these costs coincides with variational error on $\phi$; i.e., $\regret\para{\tsv{\mina}{1:T}, \tsv{\minx{{c}}}{1:T}}=   \vierr(\tsv{\mina}{1:T}, \bsetf{ \payoff(\cdot, \tsv{\maxa}{t})}_{t \in [T]})$ and $    \regret\para{\tsv{\maxa}{1:T}, \tsv{\maxx{{c}}}{1:T}} = \vierr(\tsv{\maxa}{1:T}, \bsetf{  - \payoff(\tsv{\mina}{t}, \cdot)}_{t \in [T]})$.
    Our empirical regret and generalization error bounds therefore imply
    \begin{align*}
        \vierr(\tsv{\mina}{1:T}, \bsetf{ \payoff(\cdot, \tsv{\maxa}{t})}_{t \in [T]}) \leq \lipschitz \sqrt{T} \para{4R\sqrt{2\log(1/\delta)} + 5 \sqrt{\gamma_T(\learningalgorithm_{\minactions})}} \\
        \vierr(\tsv{\maxa}{1:T}, \bsetf{  - \payoff(\tsv{\mina}{t}, \cdot)}_{t \in [T]}) \leq  \lipschitz \sqrt{T} \para{4R\sqrt{2\log(1/\delta)} + 5\sqrt{\gamma_T(\learningalgorithm_{\maxactions})}}.
    \end{align*}
    For the stated choice of $T$, $\vierr(\tsv{\mina}{1:T}, \bsetf{\payoff(\cdot, \tsv{\maxa}{t})}_{t \in [T]}) \leq T \epsilon$ and $ \vierr(\tsv{\maxa}{1:T}, \bsetf{-\payoff(\tsv{\mina}{t}, \cdot)}_{t \in [T]}) \leq T \epsilon$ with probability at least $1 - 2\delta$.
    By Fact~\ref{fact:saddlepointvi}, $\regret(\tsv{\mina}{1:T}, \bsetf{\payoff(\cdot, \tsv{\maxa}{t})}_{t \in [T]}) \leq T \epsilon$ and $\regret(\tsv{\maxa}{1:T}, \bsetf{-\payoff(\tsv{\mina}{t}, \cdot)}_{t \in [T]}) \leq T \epsilon$.
    Finally, by Fact~\ref{fact:vierrtoequilibria}, we have that $(\overline{\mina}, \overline{\maxa})$ is an $2\epsilon$-min-max equilibrium with probability at least $1 - 2\delta$.
\end{proof}

\section{Multi-Distribution Learning}
\label{subsection:formalization}

In this section, we present Algorithm~\ref{alg:collab_simple}, a general recipe for multi-distribution learning.
Algorithm~\ref{alg:collab_simple} is a general framework into which one can plug any choice of online learning algorithm to obtain a variety of multi-distribution learning guarantees.
The algorithm, which implements a form of stochastic game dynamics, uses the tools we outlined in Section~\ref{sec:tech-overview} to  reduce multi-distribution learning to the problem of solving a convex-concave game, and we then employ online learning algorithms to solve the game.
In Theorem~\ref{thm:generalconvex}, we present one example of the multi-distribution learning guarantees that Algorithm~\ref{alg:collab_deep} can provide for any online-learnable hypothesis class $\hyps$.
\begin{algorithm}[H]
	\caption{General Recipe for Multi-Distribution Learning.}
	\label{alg:collab_simple}
	\begin{algorithmic}
		\STATE \textbf{Input:} Hypothesis class $\hyps$ with parameter space $\Theta$, example oracles $\exoracle(\dist_1), \dots, \exoracle(\dist_\numdists)$, iterations $T$, online learning algorithm $\learningalgorithm_{\Theta}$ and a partial feedback online learning algorithm $\learningalgorithm_{\simplex_{\numdists\times \numlosses}}$;
		\STATE \textbf{Initialize:} $\tsv{\theta}{1} = \learningalgorithm_{\Theta}(\emptyset)$ and $\tsv{w}{1}, \tsv{I}{1} = \learningalgorithm_{\simplex_{\numdists \times \numlosses}}(\emptyset)$;
		\FOR{$\timestep = 2, \dots, T$}
        \STATE Sample $(i, j) \sim \tsv{w}{t}$ and a datapoint $\tsv{z}{t-1} \;\simiid\; \exoracle(\dist_i)$;
        \STATE Update the learner's action $\tsv{\theta}{t} = \learningalgorithm_{\Theta}(\bsetf{\theta \mapsto \inner{\nabla_{\theta} \loss_j(\hyp_\theta, \tsv{z}{\tau})}{\theta}}_{\tau \in [t-1]})$;
        \STATE For all $(i,j) \in \tsv{I}{t-1}$, sample a datapoint $\tsv{z_i}{t-1} \;\simiid\; \exoracle(\dist_i)$ for every unique $i$;
        \STATE Update the auditor's action $\tsv{w}{t}, \tsv{I}{t} = \learningalgorithm_{\simplex_{\numdists \times \numlosses}}(\bsetf{w \mapsto 1 - \sum_{i=1}^\numdists \sum_{j=1}^\numlosses w_{ij} \loss_j(\hyp_{\tsv{\theta}{\tau}}, \tsv{z_i}{\tau})}_{\tau \in [t-1]})$;
		\ENDFOR
		\STATE \textbf{Return:} $\hyp_{\overline{\theta}}$ where $\overline{\theta} = \frac{1}{T} \sum_{\timestep=1}^{T} \tsv{\theta}{t}$;
	\end{algorithmic}
\end{algorithm}

\begin{restatable}{theorem}{generalconvex}
	\label{thm:generalconvex}
    Consider a multi-distribution learning problem $(\dists, \losses, \hyps)$ with convex and $1$-smooth losses and a parameter space $\Theta$ of diameter $\diameter$: $\max_{\theta, \theta' \in \Theta} \norm{\theta - \theta'}_* \leq \diameter$.
    Let $\cQ_{\simplex_{\numdists \times \numlosses}}$ be a high-probability \cite{neu15} variant of the ELP algorithm \cite{mannor2011bandits} implemented on the partition $P = [\bset{(i, j)}_{j \in \numlosses}]_{i \in \numdists}$.
    For any choice of online learning algorithm $\cQ_\Theta$, with probability $1-\delta$, Algorithm~\ref{alg:collab_simple} returns an $\epsilon$-optimal solution $\hyp_{\overline\theta} \in \hyps$ where
    \[\epsilon \in \bigO{\sqrt{ T^{-1}\para{ \gamma_T(\learningalgorithm_{\Theta}) + \numdists\log (\numlosses \numdists / \delta) + \diameter \log(1/\delta)} }}.\]
    The sample complexity of the algorithm is $2T$.
\end{restatable}
\begin{proof}
    Algorithm~\ref{alg:collab_simple} implements Algorithm~\ref{alg:finite_general} on the convex-concave game $(\Theta, \simplex(\dists \times \losses), \phi)$, where the payoff function $\phi$ is 1-smooth and defined as $\phi(\theta, (\dist, \loss)) = \risk_{\dist, \loss}(\hyp_\theta)$.
    
    We now turn to verifying that the conditions of Lemma~\ref{lemma:zerosumasymmetricfull} are satisfied.
    Since we assume that all losses are 1-smooth in some norm $\norm{\cdot}$, the learner's payoff gradient $\nabla_{p} \phi(p, q)$ is always bounded by $1$ in the same norm, while we assume the the learner's action set diameter is at most $R$ as measured by the dual norm $\norm{\cdot}_*$.
    By linearity of expectation, we also have that the auditor's payoff gradient $\nabla_q \phi(p, q)$ is always bounded by $1$ in the infinity norm, while the auditor's action set diameter---a probability simplex---is at most $1$ as measured by the 1-norm.
    By Fact~\ref{fact:estimators}, the gradient estimators used in Algorithm~\ref{alg:finite_general}, i.e. $\theta \mapsto \inner{\nabla_{\theta} \loss_j(\hyp_\theta, \tsv{z}{\tau})}{\theta}$ and $w \mapsto 1 - \sum_{i=1}^\numdists \sum_{j=1}^\numlosses w_{ij} \loss_j(\hyp_{\tsv{\theta}{\tau}}, \tsv{z_i}{\tau})$, are unbiased, i.i.d., and $1$-bounded estimates of the payoff gradients $\nabla_{\theta} \phi(\theta, \tsv{w}{\tau})$ and $\nabla_w \phi(\tsv{\theta}{\tau}, w)$ respectively.
    Thus, all conditions of Lemma~\ref{lemma:zerosumasymmetricfull} are satisfied.
    
    We therefore know that $(\hyp_{\overline{\theta}},  \frac 1T \sum_{t=1}^T \tsv{w}{t})$ is an $\epsilon$-equilibrium with probability $1- \delta$ if
    \[
    T \geq \frac{128}{\epsilon^2}
		\para{
			\diameter^2 \log(2/ \delta) + \gamma_T (\learningalgorithm_{\Theta}) + \gamma_T (\learningalgorithm_{\simplex_{\numdists \numlosses}})
		}.
  \]
    Recalling that the regret bound of the ELP algorithm is $\gamma_T(\cQ'_{\Delta_{\numdists \numlosses}}) \in O(n \log(nm/\delta))$ (Lemma~\ref{lemma:elpconvergence}), it suffices if $
    T \geq \frac{C}{\epsilon^2}
		\para{
			\diameter^2 \log(2/ \delta) + \gamma_T (\learningalgorithm_{\Theta}) + \numdists \log(\numlosses \numdists / \delta))
		}
  $ for some universal constant $C$.
	By Fact~\ref{fact:equilsolution}, it thus follows that $\overline{\hyp} \asseq \frac 1T \sum_{t=1}^T \tsv{\hyp}{t}$ is a $2 \epsilon$-optimal solution with probability $1 - 2\delta$.

    We now resolve the sample complexity of our instantiation of Algorithm~\ref{alg:collab_simple}.
    At every timestep, the learner draws one datapoint $\tsv{z}{t}$.
    The number of datapoints that the auditor draws in any given iteration is the number of unique values of $i$ in the set $\bset{(i,j) \in \tsv{I}{t-1}}$.
    Concretely, recall that $\tsv{I}{t-1}$ denotes which cost components that the partial feedback algorithm $\cQ_{\Delta_{n \times m}}$ chooses to observe from step $t-1$, where an entry $(i,j) \in \tsv{I}{t-1}$ indicates that the auditor wishes to estimate the (in hindsight) outcome of auditing the learner on the data distribution $\dist_i$ and loss function $\loss_j$.
    We implement the ELP algorithm on the partition $P = [\bset{(1, 1), \dots, (1, \numlosses)}, \dots, \bset{(\numdists, 1), \dots, (\numdists, \numlosses)}]$, where all elements of a given group $I \in P$ correspond to the same data distribution $\dist_i$ but different choices of loss functions $\loss_j$, meaning that $\text{Unique}(\bsetf{i \mid (i, j) \in I}) = 1$.
    Since ELP guarantees that $\tsv{I}{t-1} \in P$, we can conclude the auditor only observes one datapoint per iteration.
    The total sample complexity of the algorithm is therefore $2T$.
\end{proof}
One interpretation of Theorem~\ref{thm:generalconvex} is that the worst-case sample complexity of multi-distribution learning is not significantly larger than the worst-case sample complexity of learning a single data distribution with an online-to-batch reduction.
More specifically, handling multiple data distributions and loss functions only adds an additive factor to one's sample complexity.
It may seem surprising that our sample complexity bound for multi-distribution learning---a stochastic setting---is characterized by complexity of online decision-making---an adversarial setting.
However, multi-distribution learning is inherently an online decision-making problem, as it requires one to strategize adaptively regarding the choice of data distribution to collect additional samples from.
This is in contrast to the usual single-distribution learning setting, where there is no explicit decision-making involved.

\section{Collaborative Learning}
\label{section:finiteupperbound}

In this section, we present our main result on collaborative learning: a tight bound on the sample complexity of collaborative learning in agnostic settings.
In particular, we show that the collaborative learning of a finite hypothesis class $\hyps$ on $\numdists$ data distributions requires $\Theta(\frac{\log(\setsize{\hyps}) + \numdists \log(\numdists / \delta)}{\epsilon^{2}})$ samples.
This means that, when characterizing hypothesis class complexity by $\log(\setsize{\cH})$, the worst-case sample complexity of learning $\numdists$ distributions is not significantly larger than the worst-case sample complexity of learning one data distribution.
Specifically, it requires at most a constant factor or an additive $O(\numdists \log(\numdists/\delta) / \epsilon^2)$ factor additional samples.

\subsection{Sample Complexity Upper Bound}
Theorem~\ref{thm:finite} states our sample complexity upper bound for agnostic collaborative learning.
It is a direct implication of the sample complexity of multi-distribution learning because, as we noted previously (Fact~\ref{fact:collablearning}), one can easily reduce agnostic collaborative learning to multi-distribution learning.
Theorem~\ref{thm:finite} improves over the best-known sample complexity for agnostic collaborative learning by~\citet*{nguyen_improved_2018} in two ways, giving an $\detopt+\epsilon$ bound for randomized classifiers instead of their $2 \detopt+\epsilon$ bound, and improving their sample complexity of $\bigO{\frac{1}{\epsilon^5} \para{ \log(\numdists) \log(\setsize{\hyps}) \log\para{\frac{1}{\epsilon}} + \numdists \log \para{\frac{\numdists}{\delta}}}}$ by a multiplicative factor of $\frac{1}{\epsilon^3} \log \para{\numdists} \log \para{\frac{1}{\epsilon}}$.

\begin{restatable}{theorem}{finite}
	\label{thm:finite}
    Given a set of data distributions $\dists = \bsetf{\dist_1, \dots, \dist_\numdists}$, a hypothesis class $\hyps \in \labels^\features$, and a $[0,1]$-bounded loss $\loss$, consider the collaborative learning problem $(\hyps, \dists)$.
    Consider the output ${\hyp} \in \simplex(\hyps)$ of applying Theorem~\ref{thm:generalconvex}'s algorithm to the multi-distribution learning problem $(\dists, \bset{\loss}, \simplex(\hyps))$ where the online learning algorithm $\cQ_{\simplex(\hyps)}$ is Hedge.
    With probability $1 - \delta$, $\hyp$ is an $\epsilon$-optimal solution (see \eqref{eq:deterministic_collab_learning_goal}) to $(\hyps, \dists)$ and the sample complexity is $\bigO{{\epsilon^{-2} \para{\log(\setsize{\hyps}) + \numdists \log(\numdists/\delta)}}}
	$.
\end{restatable}
\begin{proof}
    The reduction of collaborative learning to multi-distribution learning (Fact~\ref{fact:collablearning}) implies that any $\epsilon$-optimal solution to the multi-distribution learning problem $(\dists, \bset{\loss}, \simplex(\hyps))$ is an $\epsilon$-optimal solution to the collaborative learning problem $(\hyps, \dists)$.
    Fact~\ref{fact:collablearning} also establishes that  $(\dists, \bset{\loss}, \simplex(\hyps))$  has convex and 1-smooth losses, where smoothness is measured in the infinity norm.
    Since $\simplex(\hyps)$ is a probability simplex, we also have that its diameter is at most $2$ in the 1-norm.
    The guarantees of Theorem~\ref{thm:generalconvex} thus hold in our setting.
    
    Since we choose to instantiate the online learning algorithm $\cQ_{\simplex(\hyps)}$ used in Algorithm~\ref{alg:collab_simple} with Hedge, we recall that Hedge provides the regret guarantee (Lemma~\ref{lemma:egdconvergence}) of $\gamma_T(\cQ_{\simplex(\hyps)}) \in O(\log(\setsize{\hyps}))$.
    Thus, we can write the statement of Theorem~\ref{thm:generalconvex} as guaranteeing that Algorithm~\ref{alg:collab_simple} takes at most $2T$ datapoints in total and that, with probability $1 - \delta$, the output $\hyp$ of Algorithm~\ref{alg:collab_simple} is an $\epsilon$-optimal solution to $(\dists, \bset{\loss}, \simplex(\hyps))$, where $\epsilon \in \bigO{\sqrt{T^{-1} \para{\log(\setsize{\hyps}) + \numdists \log(\numdists/\delta)}}}
	$.
\end{proof}

For constants $\epsilon$ and $\delta$, our sample complexity of $\bigO{\log(\setsize{\hyps}) + \numdists \log(\numdists)}$ appears to violate the lower bound of $\Omega \para{\log(\setsize{\hyps}) \log(\numdists) + \numdists \log( \log(\setsize{\hyps}))}$ due to~\citet*{chen_tight_2018}. This discrepancy is due to a small error in the proof of that lower bound, which we have verified in private communications with the authors.

Recall that we can convert any randomized solution to a deterministic one by taking a majority vote (Fact~\ref{fact:randomizedtodet}).
Our sample complexity bound on finding randomized solutions therefore also implies a sample complexity bound on finding deterministic improper solutions to collaborative learning problems.

\begin{corollary}[Theorem~\ref{thm:finite} and Fact~\ref{fact:randomizedtodet}]
    Consider a collaborative learning problem $(\hyps, \dists)$ on a set of binary classifiers. There is an algorithm with a sample complexity of  $\bigO{{\epsilon^{-2} \para{\log(\setsize{\hyps}) + \numdists \log(\numdists/\delta)}}}
	$ that, with probability $1 - \delta$, returns a deterministic improper solution ${\hyp}_\text{Maj} \in \labels^\features$ such that $\max_{\dist \in \dists} \risk_\dist ({\hyp}_\text{Maj}) \leq 2\detopt + \epsilon$.
\end{corollary}
 
In the next subsection, we will show that this sample complexity upper bound is tight up to double-log factors and exactly tight in the regime where $\numdists \in O(\log(\setsize{\hyps})$.

\subsection{Sample Complexity Lower Bound}
\label{sec:mainlowerbound}

\def\probinstance{\mathbb{P}}
\def\instanceset{\mathbb{V}}
\def\instance{V}

We now provide matching lower bounds on the sample complexity of agnostic collaborative learning.
We note that these lower bounds hold for any collaborative learning algorithm that returns $\epsilon$-optimal solutions, regardless of whether those algorithms perform sampling on-demand and regardless of whether the algorithms return randomized or deterministic and proper or improper solutions. 
We also note that the data distributions we construct to establish these lower bounds are not exotic.
For example, to prove these lower bounds, we construct a set of data distributions where all data distributions share the exact same feature distribution and all but one distribution share the exact same label distribution.

Theorem~\ref{theorem:lowerboundagnostic} states our lower bound.
In this theorem, we refer to an algorithm as an $(\epsilon, \delta)$-collaborative learning algorithm if, for every collaborative learning problem $(\hyps, \dists)$, the algorithm returns an $\epsilon$-optimal solution $\hyp$, i.e., satisfying Equation~\ref{eq:deterministic_collab_learning_goal}, with probability at least $1-\delta$.
We say that a learning algorithm $\learningalgorithm{}$ is an $(\epsilon,\delta)$-optimal collaborative learning algorithm for a specific set of collaborative learning problems $\instanceset \asseq \bset{(\hyps_i, \dists_i)}_i$ if, given any problem $(\hyps_i, \dists_i) \in \instanceset$, with probability at least $1 - \delta$ the output of $\learningalgorithm{}$ is an $\epsilon$-optimal solution.

\begin{restatable}{theorem}{lowerboundagnostic}
	\label{theorem:lowerboundagnostic}
	Take any $\numdists, d \in \integers_+$, $\epsilon, \delta \in (0, 1/8)$, and $(\epsilon, \delta)$-collaborative learning algorithm $\learningalgorithm$. There exists a collaborative learning problem $(\hyps, \dists)$ with $\setsize{\dists}= \numdists$ and $\setsize{\hyps}=2^d$, on which $\learningalgorithm$ takes at least $\Omega \para{\frac{1}{\epsilon^2} \para{d + \numdists \log(\min\bsetf{\numdists, d} / \delta)}}$ samples.
    When $\numdists \leq d$, this lower bound becomes $\Omega \para{\frac{1}{\epsilon^2} \para{d + \numdists \log(\numdists / \delta)}}$. 
\end{restatable}

Before we proceed to a proof of this theorem, we first define a notion of expected sample complexity.
Take any collaborative learning problem $\instance = (\hyps, \dists)$; we use $N_\learningalgorithm(\instance)$ to denote the expected sample complexity of a collaborative learning algorithm $\learningalgorithm$ on the problem $\instance$, where the expectation is taken both over the randomness of data samples and the algorithm's randomness.
Similarly, given a probability distribution  $\probinstance$  over a set of collaborative learning problems $\instanceset \asseq \bset{(\hyps_i, \dists_i)}_i$, we define expected sample complexity as $N_{\learningalgorithm}(\probinstance) = \EEs{\instance \sim \probinstance}{N_\learningalgorithm(\instance)}$.

We now prove two lemmas, Lemma \ref{lemma:agnostic} and Lemma \ref{lemma:multi}, that directly imply Theorem \ref{theorem:lowerboundagnostic}.
Lemma~\ref{lemma:agnostic} is a standard lower bound on the sample complexity of agnostic PAC learning, and provides the unsurprising $\Omega(\frac{d}{\epsilon^2})$ lower bound summand.
Lemma~\ref{lemma:multi} is more involved and provides the $\Omega(\frac{n \log(\min\bset{n, d}/\delta)}{\epsilon^2})$ summand in our lower bound.
\begin{lemma}
\label{lemma:agnostic}
Take any $\numdists, d \in \integers_+$, $\epsilon, \delta \in (0, 1/8)$, and collaborative learning algorithm $\learningalgorithm$. There exists a set of collaborative learning problems $\instanceset$ on which, if $\learningalgorithm$ is $(\epsilon, \delta)$-optimal, $\learningalgorithm$ takes at least $\Omega \para{\frac{\log \setsize{\hyps} }{\epsilon^2}}$ samples and where, for every $(\hyps, \dists) \in \instanceset$, $\setsize{\dists}= \numdists$ and $\setsize{\hyps}=2^d$.
\end{lemma}
\begin{proof}
This claim follows directly from the  standard lower bound on sample complexity of agnostic probably-approximately-correct (PAC) learning~\cite{valiant_theory_1984}, since we can reduce any single-distribution learning problem to multi-distribution learning problem by defining multiple copies of a data distribution.
We defer interested readers to \citet{ehrenfeucht_general_1989}.
\end{proof}

\begin{lemma}
\label{lemma:multi}
Take any $\numdists, d \in \integers_+$, $\epsilon, \delta \in (0, 1/8)$, and $(\epsilon, \delta)$-collaborative learning algorithm $\learningalgorithm$.
There exists a set of collaborative learning problems $\instanceset$ on which $\learningalgorithm$ takes at least $\Omega \para{\frac{1}{\epsilon^2} \para{\numdists \log(k / \delta)}}$ samples and where, for every $(\hyps, \dists) \in \instanceset$, $\setsize{\dists}= \numdists$ and $\setsize{\hyps}=2^{d}$ with $k \asseq \min \bset{\numdists, d}$.
\end{lemma}
\begin{proof}
We prove this lower bound constructively by defining multiple sets of collaborative learning instances: $\bsetf{\instanceset_{w \eta, w}}_{w,\eta \in \mathbb{N}}$.
At a high-level, the proof of this lower bound will follow from proving that multi-distribution learning allows one to solve multiple single-distribution learning problems simultaneously with constant probability using a boosting-like algorithm.

We now detail our fairly technical construction of these collaborative learning instances. For every set of instances $\instanceset_{u, w}$, we require all instances $(\hyps, \dists) \in \instanceset_{u, w}$ to share a feature space $\features = \bset{1, \dots, w}$, label space $\labels = \bsetf{\pm 1}$, hypothesis class $\hyps = \labels^\features$, and 0/1 loss $\loss$.
For every $x \in [w]$ and $y \in \bsetf{\pm 1}$, we define distributions $\dist_{x}$ and $\dist_{x}'$ as having the probability mass functions $\Pr_{\dist_{x}} \para{{x, y}} =  \frac{1}{2} - 2y\epsilon$ and $\Pr_{\dist_{x}'} \para{{x, y}} = \frac{1 }{2} + 4y\epsilon$.
Let $\dists_- = \bigcup_{x \in [w]} \bsetf{\dist_{x}}^\eta$ be an ordered list of distributions, and for every $x \in [w]$ and $i \in [\eta]$, define $\dists_{x,i}$ to be a set of distributions identical to $\dists_-$ except with the $i$th copy of distribution $\dist_x$ replaced with distribution $\dist_{x}'$.
Let $\probinstance_{\eta w,w}$ be a distribution over collaborative learning instances that, with probability $\frac 12$, returns $(\hyps, \dists_-)$ and for every $i \in [\eta], x^* \in [w]$, with probability $\frac {1}{2 w \eta}$ returns $(\hyps, \dists_{x^*, i})$.
Observe that $\probinstance_{\eta w,w}$ is a distribution over collaborative learning problems where $\setsize{\hyps} = 2^w$ and $\setsize{\dists} = u$.
The following claims characterize sample complexity lower bounds on $\probinstance_{u,w}$.

\begin{claim}
\label{claim:reductionlowerbound}
Consider any $\epsilon \in (0, 1/2)$, $\delta \in (0,1)$, and collaborative learning algorithm $\learningalgorithm$ that is $(\epsilon, \delta)$-optimal for $\instanceset_{\eta, 1}$.
The expected sample complexity of $\learningalgorithm$ is at least $\frac{\eta}{256 \epsilon^2} \log(1/2\delta)$.
\end{claim}

\begin{claim}
\label{claim:reductionupperbound}
Consider any $\epsilon \in (0, 1/2)$ and $\delta \in (0,1)$.
Suppose there exists a collaborative learning algorithm $\learningalgorithm$ that is $(\epsilon, \delta)$-optimal for $\instanceset_{\eta w,w}$ and has an expected sample complexity of $N$ under $\probinstance_{\eta w,w}$.
Then there exists an $(\epsilon, \frac{8 \delta}{7w})$-learning algorithm $\learningalgorithm'$ for $\instanceset_{\eta, 1}$ under $\probinstance_{\eta, 1}$ with an expected sample complexity on $\probinstance_{\eta, 1}$ of $\frac{8}{7 w} N$.
\end{claim}

Since our desired lower bound is weakly monotonic in $n, d$, we fix the smallest choice of $\eta, d \in \integers_+$ and $\epsilon, \delta \in (0, 1/8)$ such that $n = \eta \cdot d$.
Combining claims \ref{claim:reductionlowerbound} and \ref{claim:reductionupperbound}, we see that any $(\epsilon,\delta)$ collaborative learning algorithm $\learningalgorithm$ for $\instanceset_{\numdists, d}$ has an expected sample complexity on $\probinstance_{\numdists, d}$ of at least $N \geq \frac{7 \numdists}{2048 \epsilon^2} \log\para{\frac{7 d}{16 \delta}}$.
By the probabilistic method, for at least some collaborative learning problem in the set $\instanceset_{\numdists, d}$, our learning algorithm $\learningalgorithm$ must have a sample complexity of $\Omega\para{\frac{7 \numdists}{2048 \epsilon^2} \log\para{\frac{7 d}{16 \delta}}}$.
\end{proof}

\begin{proof}[Proof of Claim \ref{claim:reductionlowerbound}]
Consider $\eta$ two-sided coins.
Under a $H_0$ hypothesis, all coins are biased towards tails with probability $1/2+2\epsilon$.
Under a $H_i$ hypothesis, the $i$th coin is biased towards heads with probability $1/2+4\epsilon$.
Let $\Pr$ be a probability distribution on $H \in \bset{H_i}_{i=0}^\eta$ with $\Pr(H_0) = 1/2$ and $\Pr(H_1) = \dots = \Pr(H_\eta) = \frac{1}{2\eta}$.
Given an $(\epsilon,\delta)$-algorithm $\learningalgorithm$ for $\instanceset_{\eta,1}$ with an expected sample complexity of $N$ (under $\probinstance_{\eta,1}$), we can construct a coin algorithm $\learningalgorithm'$ with an expected sample complexity of $N$ (under $\Pr$) and that, under any hypothesis, with probability at least $1 - \delta$, can identify whether $H_0$ is false.

To see this, have $\learningalgorithm'$ run $\learningalgorithm$ by simulating draws from the $i$th distribution by flipping the $i$th coin.
If all coins are biased towards tails with probability $1/2+2\epsilon$, any $\epsilon$-error hypothesis $\hyp$ must satisfy $\Pr(h(1) = +) > 1/2$.
Conversely, if one coin is biased towards heads, any $\epsilon$-error hypothesis $\hyp$ must satisfy $\Pr(h(1) = +) < 1/2$.

Suppose $\learningalgorithm'$, conditioned on $H_0$, correctly predicts $H_0$ with probability at least $1 - \delta$.
Then, suppose $\learningalgorithm'$, under $H_0$, takes no more than $T_i$ flips from the $i$th coin.
Let $p_{i,j_1:j_2}$ be a probability distribution over $\bset{0,1}$ corresponding to the outcomes of the $j_1$st to $j_2$nd coin toss by $\learningalgorithm'$ under $H_i$.
Let $p^*_j$ be a uniform distribution over $\bset{0,1}^j$.
Since $p_{i,j:j}$ and $p^*_j$ are Bernoulli distributions with a parameter within $4 \epsilon$ of $1/2$, for $\epsilon < 1/2$, $\text{KL}(p_{i,j:j}, p^*_1) < 128 \epsilon^2$ \cite{zhang_information-theoretic_2019}.
Moreover, $\text{KL}(p_{i,1:j}, p^*_j) < 128 j \epsilon^2$ by tensorization and $\text{TV}(p_{i,1:j}, p^*_j) \leq 8 \epsilon \sqrt{j}$ by Pinsker's inequality.
Let $E$ be the set of outcomes of $T_i$ flips under which $\learningalgorithm'$ predicts $H_0$.
By correctness under $H_0$, we have that $\Pr_{H_0}(E) \geq 1 - \delta$.
Thus, total variation distance implies $1 - \delta - 8 \epsilon \sqrt{j} < \Pr_{H_i}(E)$.
Since $\Pr_{H_i}(E) < \delta$, we have that $\frac{1}{128 \epsilon^2} \para{1 - 2 \delta}^2 < T_i$.
Thus, if $\learningalgorithm'$ is $\delta$ accurate under all hypotheses, under $H_0$, $\learningalgorithm'$ must take at least $\frac{\eta}{128 \epsilon^2} \para{1 - 2 \delta}^2 < \frac{\eta}{128 \epsilon^2} \log(1/2\delta)$ samples from each distribution.
Thus, the expected sample complexity of $\learningalgorithm'$---and similarly that of $\learningalgorithm$ under $\probinstance_{\eta, 1}$---must be at least $\frac{\eta}{256 \epsilon^2} \log(1/2\delta)$.
\end{proof}

\begin{proof}[Proof of Claim \ref{claim:reductionupperbound}]
This claim is similar to the lower bounds of~\citet{blum_collaborative_2017} and~\citet{karp_noisy_2007}.
We construct $\learningalgorithm'$ as follows.
Define the shorthand $I_j \asseq [(j-1) \eta + 1, j \eta]$.
Consider any problem $\instance' = (\hyps, \dists) \in \instanceset_{\eta, 1}$.
\begin{enumerate}
    \item $\learningalgorithm'$ draws an imaginary problem $(\hyps, \dists') \in \instanceset_{\eta w, w}$ and chooses an index $i \in [w]$ uniformly at random.
    \item $\learningalgorithm'$ simulates algorithm $\learningalgorithm$ on $(\hyps, \dists')$: when $\learningalgorithm$ tries to sample a datapoint from distribution $\dist'_j$ where $j \notin I_i$, return a sample from $\dist'_j$; when $j \in I_i$, return a sampled datapoint from $\dist_{j - (i-1)\eta}$. 
    \item When $\learningalgorithm$ terminates and returns a classifier $\hyp$, $\learningalgorithm'$ checks whether, for every $j \neq i$: $\max_{r \in I_j} \risk_{\dist_r}(\hyp)
        < \frac{1}{2}$.
    If this condition is satisfied, $\learningalgorithm'$ returns $\hyp(1) = \hyp(i)$.
    If not, we repeat from Step 1. We denote the number of total iterations by $T$.
\end{enumerate}

Consider the probability $p_i$ that, in the third step, for every $j \neq i$ we have $\max_{r \in I_j} \risk_{\dist_r}(\hyp)
        < \frac{1}{2}$ but $\max_{r \in I_i} \risk_{\dist_r}(\hyp)
        \geq \frac{1}{2}$.
Let $E_\timestep$ denote the event that $\learningalgorithm'$ returns an at least $\epsilon$-error hypothesis after $\timestep$ iterations of our procedure.
Noting that $E_\timestep$ can only occur if $\learningalgorithm$ failed all $\timestep-1$ iterations before and at the $\timestep$th iteration, Step 3 fails to catch the bad hypothesis for $\dist_i$.
By assumption, $\delta \geq \sum_{i=1}^{w} p_i$. By symmetry of our construction $\instanceset$ and recalling $\delta < 1/8$:
$
    \sum_{\timestep=1}^\infty \Pr\para{
    E_\timestep
    }
    \leq \sum_{\timestep=1}^\infty \delta^{\timestep-1} \frac{1}{w} \sum_{i=1}^w p_i \leq \sum_{\timestep=1}^\infty \delta^\timestep / w
    \leq \frac{8 \delta}{7 w}$.
Thus, $\learningalgorithm'$ is an $(\epsilon,\frac{8\delta}{7w})$-algorithm for $\probinstance_{\eta,1}$.

We now bound the sample complexity of $\learningalgorithm'$.
Let $N_{\learningalgorithm'}(\timestep)$ denote the number of samples that $\learningalgorithm'$ takes from $\instance'$ on the $\timestep$th iteration.
Note that $N_{\learningalgorithm'}(1), N_{\learningalgorithm'}(2), \dots$ are i.i.d.
In addition, by the symmetry of $\instanceset$ and linearity of expectation, $\EEs{\instance' \in \probinstance_{\eta,1}}{N_{\learningalgorithm'}(\timestep)} = m/w$.
Thus, $
     \EEs{{\instance'}}{\sum_{\timestep=1}^T N_{\learningalgorithm'}(\timestep)}
     =
     \EEs{{\instance'}}{T}
     \EEs{{\instance'}}{N_{\learningalgorithm'}(1)}
     =  \EEs{{\instance'}}{T} m/w$.
We can upper bound $T$ by observing that our procedure only repeats if $\learningalgorithm$ fails: $\EEs{{\instance'}}{T}
    = \sum_{\timestep=1}^\infty \Pr(T \geq \timestep)
    \leq \sum_{\timestep=0}^\infty \delta^\timestep \leq \frac{1}{1-\delta} \leq \frac{8}{7}$.
Thus, $\learningalgorithm'$ has an expected sample complexity of at most $\frac{8m}{7w}$.
\end{proof}

\section{Group DRO and Agnostic Federated Learning}
\label{section:gdroupperbound}

In this section, we present our main result on the sample complexity of the group distributionally robust optimization framework
of~\citet{sagawa_distributionally_2020} and the agnostic federated learning framework of~\citet{mohri_agnostic_2019}.
We show that the worst-case sample complexity of group DRO, and equivalently agnostic federated learning, is greater than that of online convex optimization by only a constant factor and an additive $O(\numdists \log(\numdists/\delta)/\epsilon^2)$ samples.
This sample complexity upper bound is tight for a difficult class of problems---a class that coincides with collaborative learning.
Since the settings of group DRO and agnostic federated learning are generally equivalent, we state the results explicitly for group DRO, with the understanding that the same results apply to agnostic federated learning.

\paragraph{Setup.}
Group distributionally robust optimization is typically studied in a convex optimization setting where the hypothesis class is parameterized by a convex compact parameter class and the loss function is smooth and convex in the parameterization.
As noted previously, this means that the group DRO setting coincides with general setting of multi-distribution learning with a single smooth convex loss.
Moreover, group DRO is usually formulated in a setting where the parameter space admits mirror descent approaches.

We first present the definitions which are necessary for describing mirror descent guarantees.
A distance-generating function on a parameter space $\Theta$ is a continuous and strongly convex, modulus 1, function $\dgf: \Theta \rightarrow \reals$, where there exists a non-empty subset of the parameter space $\convexsetcore \subset \Theta$ where the subdifferential $\partial \dgf$ is non-empty and $\partial \dgf$ admits a continuous selection on $\convexsetcore$.
The center of $\Theta$ with respect to $\dgf$ is denoted as $\theta^c \asseq \argmin_{\theta \in \convexsetcore} \dgf(\theta)$.
The Prox function (Bregman divergence) $\proxfunction: \convexsetcore \times \convexset \rightarrow \reals^+$ associated with a distance-generating function $\dgf: \convexset \rightarrow \reals$ is defined as $\proxfunction(w, u) \asseq \dgf(u) - \dgf(w) - \innerproduct{\dgf'(w)}{u-w}$.
Bregman radius, which is a measure for how difficult it is to learn a parameter class, is then defined as follows.
\begin{definition}
	\label{def:bregmanradius}
	Given a convex set $\Theta$ with a distance-generating function $\dgf$, the Bregman radius is defined as $D_\Theta \asseq \max_{u \in \Theta} \proxfunction(\theta^c, u)$ where $\theta^c$ is the center of $\Theta$.
\end{definition}
\noindent
A bounded Bregman radius allows one to apply online mirror descent~\cite{beck_mirror_2003} as an online learning algorithm, with a regret guarantee of $\gamma_T(\cQ_\Theta) \leq D_\Theta$.
In group DRO, $D_\Theta$ is typically assumed to be small.

\paragraph{Sample complexity upper bound.}
Theorem~\ref{thm:gdro} states our sample complexity bound for group distributionally robust optimization.
This sample complexity bound is a direct implication of our multi-distribution learning sample complexity bound.
This theorem establishes the first generalization bound for the problem of group distributionally robust optimization~\cite{sagawa_distributionally_2020} and improves, by a factor of $\numdists$, existing sample complexity bounds for agnostic federated learning
\cite{mohri_agnostic_2019}.
This significant improvement in sample complexity over \citet{mohri_agnostic_2019} is attained by sampling data on-demand, whereas~\citet{mohri_agnostic_2019} work with a distribution over groups/clients that is fixed a priori.
\begin{restatable}{theorem}{gdro}
	\label{thm:gdro}
Given a set of data distributions $\dists = \bsetf{\dist_1, \dots, \dist_\numdists}$, a hypothesis class $\hyps$ with a Bregman radius of $\dist_\Theta$ and a diameter of $R$, and a $1$-smooth loss $\loss$, consider the group distributionally robust optimization problem $(\dists, \bset{\loss}, \hyps)$.
Consider the output $\theta \in \Theta$ arising from applying Theorem~\ref{thm:generalconvex}'s algorithm, choosing the online learning algorithm $\cQ$ to be online mirror descent.
 With probability $1 - \delta$, $\hyp$ is an $\epsilon$-optimal solution (see \eqref{eq:dro}) and the sample complexity is $\bigO{{\epsilon^{-2} \para{\dist_\Theta + \numdists \log(\numdists / \delta) + R \log(1/\delta)}}}
	$.
\end{restatable}
\begin{proof}
    This claim follows directly by Theorem~\ref{thm:generalconvex} since group distributionally robust optimization is equivalent to multi-distribution learning on a single smooth convex loss.
    Recall that, for a convex parameter space with Bregman radius $\dist_\Theta$ for a distance-generating function $\dgf$, running the online mirror descent algorithm with respect to $\dgf$ guarantees a regret bound of $\gamma_T(\learningalgorithm_\Theta) \leq \dist_\Theta$~\cite{beck_mirror_2003}.
    We directly plug this online convex optimization regret bound into Theorem~\ref{thm:generalconvex}.
\end{proof}
\noindent
This sample complexity bound for finding a group DRO solution with low expected loss also trivially implies a bound on the number of mirror descent iterations that are necessary to find a group DRO or agnostic federated learning solution with low empirical training error.
This question was considered by~\citet{sagawa_distributionally_2020} who presented an iteration complexity bound that we improve upon by a factor of  $\numdists$.

\begin{corollary}[Theorem~\ref{thm:gdro}]
	\label{corr:gdro}
  Consider a group distributionally robust optimization problem $(\dists, \bset{\loss}, \hyps)$.
	For every $\dist \in \dists$, let ${B}_\dist \sim \dist$ be a non-empty batch of i.i.d.\ datapoints and $\dist'$ be the empirical distribution of ${B}_\dist$.
 There is an algorithm that only requires  $\bigO{\epsilon^{-2} \para{\dist_\Theta + \numdists \log( \numdists / \delta) + R \log(1/\delta)}}$ iterations of mirror descent steps to output, with probability $1-\delta$,  an \emph{empirically} $\epsilon$-optimal solution.
\end{corollary}

\paragraph{Sample complexity lower bound.}
There exists a class of difficult group distributionally robust optimization problems for which our stated sample complexity upper bounds are tight.
This is because we can reduce any collaborative learning problem to multi-distribution learning with a single smooth convex loss, and equivalently, group DRO.
Thus, our sample complexity lower bound for collaborative learning directly implies a lower bound for group DRO for a class of difficult cases.
We formally state this corollary of Theorem~\ref{theorem:lowerboundagnostic} below.
\begin{restatable}{corollary}{lowerboundagnosticfull}
\label{corollary:lowerboundagnosticfull}
Take any $\numdists, m \in \mathbb{N}$ and $\epsilon, \delta \in (0, 1/8)$.
There exists a finite set $\instanceset$ of group distributionally robust optimization problems with $1$-smooth losses and parameter spaces of unit diameter and finite Bregman radius $\dist_\Theta$, where every $(\epsilon, \delta)$-algorithm $\learningalgorithm$ has a sample complexity in $ \Omega\para{\frac{\dist_\Theta + \numdists \log(\min\bset{\numdists, \dist_\Theta} / \delta)}{\epsilon^2}}$.
\end{restatable}

\section{Extensions to Infinite Classes of Binary Classifiers}
In this section, we study the sample complexity of multi-distribution learning when the hypothesis class is infinite but combinatorially bounded.
In particular, we will study multi-distribution learning problems involving binary classification tasks and hypothesis classes of finite VC dimension or finite Littlestone dimension \cite{littlestone_learning_1987}.
For succinctness, we state all results in this section for the collaborative learning setting, but note that these results extend readily to the general multi-distribution learning setting.

\paragraph{Littlestone dimension.}
The Littlestone dimension of a set of binary classifiers quantifies the set's online learnability \cite{littlestone_learning_1987}.
Formally, consider a supervised learning setting with domain $\cX$ and a set of binary classifiers $\cH$.
Consider a full binary tree of depth $d$, such that each node in the tree is labeled by a feature $x \in \cX$.
We say the tree is shattered by $\cH$ if for every set of labels $\bset{y_i}_{i=1}^d \in \bset{\pm1}^d$, the root-to-leaf path $x_1, \dots, x_d$ that is defined by starting at the root and moving to the left child if $y_i = +1$ and right if $y_i = -1$, there exists a classifier $\hyp \in \hyps$ such that $h(x_i) =y_i$ for all $i \in [d]$---that is, $\hyp$ agrees with the labels we used to reach nodes in the path.
In other words, a tree is shattered if every path in the tree is labeled by some hypothesis $h \in \cH$.
We say that the Littlestone dimension of the classifiers $\cH$ is $d$ if $d$ is the maximal depth of a tree that is shattered by $\hyps$.

It is not hard to see that Littlestone dimension upper bounds VC dimension and lower bounds log-cardinality $\log(\setsize{\cH})$.
For binary classifier multi-distribution learning problems, we can strengthen our collaborative learning sample complexity upper bound of Theorem~\ref{thm:finite} to be stated in terms of the Littlestone dimension of a hypothesis class $\lsdt \para{\hyps}$ rather than $\log(\setsize{\cH})$.
This is because there exists an online learning algorithm that guarantees a regret bound of $\gamma_T(\cQ_{\simplex(\hyps)}) \in O(\lsdt(\hyps))$ that we can have the learner play instead of an algorithm like Hedge.
\begin{restatable}[Littlestone Dimension Variant of Theorem~\ref{thm:finite}]{theorem}{littlestone}
	\label{thm:littlestone}
    Given a set of data distributions $\dists = \bset{D_1, \dots, D_\numdists}$, a hypothesis class of binary classifiers $\hyps \in \bset{0, 1}^\cX$, and a $[0,1]$-bounded loss $\loss$, consider the collaborative learning problem $(\hyps, \dists)$.
    Consider the output $\hyp \in \simplex(\hyps)$ of applying Theorem~\ref{thm:generalconvex}'s algorithm to the multi-distribution learning problem $(\dists, \bset{\loss}, \simplex(\hyps))$ where the online learning algorithm $\cQ_{\simplex(\hyps)}$ is the agnostic Standard-Optimal-Algorithm of \citet{alon_adversarial_2021}. With probability $1 - \delta$, $\hyp$ is an $\epsilon$-optimal solution (see \eqref{eq:deterministic_collab_learning_goal}) to $(\hyps, \dists)$ and the sample complexity is $\bigO{{\epsilon^{-2} \para{\lsdt \para{\hyps} + \numdists \log(\numdists/\delta)}}}$.
\end{restatable}
\begin{proof}
    By Fact~\ref{fact:collablearning}, we can reduce the collaborative learning problem $(\hyps, \dists)$ to solving the multi-distribution learning problem $(\dists, \bset{\loss}, \simplex(\hyps))$
    The agnostic SOA algorithm of \citet{alon_adversarial_2021} guarantees a regret bound of $\gamma_T(\learningalgorithm_{\simplex(\hyps)}) = \lsdt(\hyps)$.
    Our claim therefore follows by Theorem~\ref{thm:generalconvex}.
\end{proof}
\noindent
We remark that a similar sample complexity bound can be achieved using the original Standard Optimal Algorithm (SOA) of~\citet{littlestone_learning_1987} instead of the implicit algorithm of~\citet{alon_adversarial_2021}, as SOA guarantees a regret bound of $\gamma_T(\cQ_{\simplex(\hyps)}) \in \bigO{\sqrt{\lsdt(\hyps) T \log(T)}}$.

\paragraph{VC dimension.}
It is also nature to ask for the sample complexity of multi-distribution learning in terms of  VC dimension $\vcdt(\hyps)$, which characterizes the sample complexity of learning a single data distribution.
For example, \citet{blum_collaborative_2017,nguyen_improved_2018,chen_tight_2018} provided upper bounds for binary classification multi-distribution learning that are identical to their upper bounds in Table \ref{tab:summary} but replacing $\log (\setsize{\hyps})$ with $\vcdt(\hyps)$.
We now show a similar result to Theorem~\ref{thm:finite} also holds with dependence on the VC dimension of $\hyps$ only when additional mild assumptions hold.
In particular, one can run Algorithm~\ref{alg:collab_simple} on a hypothesis class $\hyps'$ that is known to be an $\epsilon$-net of $\hyps$ with respect to each distribution in $\dists$.
Such an $\epsilon$-net of size $(\numdists /\epsilon)^{\bigO{\vcdt(\hyps)}}$ necessarily exists (see, e.g.,~\citep{anthony_neural_2002}).
For example, we can project $\hyps$ onto the union of datapoints sampled from each distribution $\dist \in \dists$.
When such a $\hyps'$ is known in advance, we may directly run Algorithm~\ref{alg:collab_simple} with $\hyps'$.
\begin{restatable}{corollary}{coveringtheorem}
    \label{corollary:covering}
    Given a set of data distributions $\dists = \bset{D_1, \dots, D_\numdists}$, a hypothesis class of binary classifiers $\hyps \in \bset{0, 1}^\cX$ of VC dimension $d$, and a $[0,1]$-bounded loss $\loss$, consider the collaborative learning problem $(\hyps, \dists)$.
    Suppose we are further given a set of classifiers of size $\mathrm{poly} \para{(\numdists/\epsilon)^{d}, \epsilon, d, \numdists}$ that is an $\epsilon$-net of $\hyps$ for each distribution $\dist \in \dists$.
    There is an algorithm that, with probability $1-\delta$, returns an $\epsilon$-optimal solution (see \eqref{eq:deterministic_collab_learning_goal}) to $(\hyps, \dists)$ with a sample complexity of $\bigO{{\epsilon^{-2} \para{d \log(d \numdists /\epsilon) + \numdists \log(\numdists/\delta)}}}$.
\end{restatable}

It is not strictly necessary to know an $\epsilon$-net in advance. Instead, one can compute a net from samples or from other information about distributions in $\dists$.
There a range of assumptions that allow us to compute such an $\epsilon$-net from samples, without incurring a significant increase in sample complexity.
For example, when $\epsilon$ is sufficiently small, specifically $\epsilon \in \bigO{1/\numdists}$ (\emph{Assumption 1}), taking an $\epsilon$-net only increases the sample complexity bound by constant factors versus knowing an $\epsilon$-net in advance.
Additional examples include:
\begin{itemize}
    \item \emph{Assumption 2}: we know the marginal distribution for all $\dist \in \dists$;
    \item \emph{Assumption 3}: we have access to $\numdists$ marginal distributions $P_1, \dots, P_\numdists$ such that for all $x \in \features$, $\dist_i(A) \leq p_i(A)\mathrm{poly}(1/\epsilon, \vcd(\hyps), \numdists)$ for all $A\subseteq \features$, where $p_i$ and $\dist_i$ are the densities of $P_i$ and $\dist_i$, respectively.
\end{itemize}
These latter two assumptions allow one to construct $\epsilon$-nets of small size for free. 
\begin{restatable}{theorem}{vcdim}
	\label{thm:vcdim}
 
    Given a set of data distributions $\dists = \bset{D_1, \dots, D_\numdists}$, a hypothesis class of binary classifiers $\hyps \in \bset{0, 1}^\cX$ of VC dimension $d$, and a $[0,1]$-bounded loss $\loss$, consider the collaborative learning problem $(\hyps, \dists)$.
    If any of Assumptions 1, 2 or 3 is met, there is an algorithm that, with probability $1-\delta$, returns an $\epsilon$-optimal solution (see \eqref{eq:deterministic_collab_learning_goal}) to $(\hyps, \dists)$ with a sample complexity of $
		\bigO{
			\epsilon^{-2} \para{d\log(dn/\epsilon) + \numdists \log( \numdists/ \delta)}
		}
	$.
\end{restatable}

\begin{proof}

For a data distribution $\dist$, we will use $\dist_\features$ to denote the marginal distribution of $\dist$.
We also use the shorthand $d_\infty(P || Q) \asseq \sup_{x \in \features_Q} \frac{P(x)}{Q(x)}$, where $d_\alpha(P || Q) \asseq 2^{\dist_\alpha(P || Q)}$ can be understood as the power of the Renyi divergence $\dist_\alpha(P || Q)$.
We first recall a standard fact about covering with projections.
\begin{lemma}[Corollary 3.7 in~\citet{hausslerEpsilonNets1986}]
    \label{lemma:net}
    Let $\cF$ be a function class consisting of functions from $\cX$ to $[0, 1]$ and let $\cP$ be a probability measure on $\cX$.
    Given $N \geq \frac{8d}{\epsilon} \log \frac{8d}{\epsilon} + \frac{4}{\epsilon} \log \frac{2}{\delta}$ independent samples $\bx$ from $\cP$, with probability at least $1 - \delta$, the projection of $\cF$ on $\bx$ constitutes an $\epsilon$-net. That is, for any $f_1, f_2 \in \cF$ where $\Pr_{x \sim \cP}(f_1(x) \neq f_2(x)) \geq \epsilon$, $\norm{f_1(x) - f_2(x)}_{\bx} > 0$.
\end{lemma}

The following corollaries of Theorem~\ref{thm:finite} directly imply Theorem \ref{thm:vcdim}.
\begin{restatable}[Assumption 1]{corollary}{coveringcorr}
    \label{corollary:covernaive}
    For $\epsilon \in \bigO{ 1/\numdists}$, there is an algorithm that, with probability $1-\delta$, returns an  $\epsilon$-optimal solution $\overline{\hyp} \in \simplex(\hyps)$  
    using a number of samples that is
$
	    \bigO{
	    \frac{d\log(dn/\epsilon) + \numdists \log( \numdists / \delta)}{\epsilon^2}
	    }.
 $
\end{restatable}
\begin{proof}
By Lemma~\ref{lemma:net}, sampling $\bigO{\frac{nd}{\epsilon} \log(\frac{d}{\epsilon}) + \frac{n}{\epsilon} \log(\frac{n}{\epsilon})}$ datapoints provides a covering of $\hyps$ is that is simultaneously an $\epsilon$-net for every $\dist \in \dists$ with probability at least $1 - \delta$.
Moreover, by the Sauer-Shelah lemma, this net is of size $\bigO{\para{\frac{\log(dn/\epsilon) + n \log(n / \delta)}{\epsilon^2}}^d}$.
The claim then follows from Corollary~\ref{corollary:covering}, noting that since $\epsilon \in \bigO{1/n}$, we only needed to sample an additional $\bigO{\frac{d}{\epsilon^2} \log(\frac{d}{\epsilon}) + \frac{n}{\epsilon} \log(\frac{n}{\epsilon})} \subset \bigO{\frac{nd}{\epsilon} \log(\frac{d}{\epsilon}) + \frac{n}{\epsilon} \log(\frac{n}{\epsilon})}$ datapoints to form the cover.
\end{proof}

\begin{restatable}[Assumption 2]{corollary}{covering}
    \label{corollary:coveringweak}
    We say an algorithm has weak unlabeled access if the algorithm can access, for each $\dist \in \dists$, a marginal distribution $\dist'_\features$ such that $\dist_\infty(\dist'_\features || \dist_\features) \in \mathrm{poly}(1/\epsilon, d,n)$, with probability $1-\delta$.
    There is an algorithm that, given weak access, with probability $1-\delta$, returns an  $\epsilon$-optimal solution $\overline{\hyp} \in \simplex(\hyps)$  
    using a number of samples that is
$
	    \bigO{
	    \frac{d\log(dn/\epsilon) + \numdists \log( \numdists / \delta)}{\epsilon^2}
	    }.
 $
\end{restatable}
\begin{proof}
Observe that when $\dist_\infty(\dist'_\features || \dist_\features) < \gamma$, $\dist'_\features$ can be written as a mixture over $\dist_\features$ with probability at least $\frac{1}{\gamma}$ and some other distribution $\tilde{\dist}_\features$ with probability at most $1 - \frac{1}{\gamma}$.
Once again invoking uniform convergence, we observe that sampling $\Theta \para{\dist_\infty(\dist'_\features || \dist_\features) \frac{d \log(d / \epsilon) + \log(1/\delta)}{\epsilon^2}}$ i.i.d.\ samples from distribution $\dist_\features'$, with probability at least $1 - \delta$, yields an $\epsilon$-covering on $\dist$.
By the Sauer-Shelah lemma, the resulting covering $\hyps'_{\dist}$ is of size $\bigO{(\mathrm{poly}(1/\epsilon, d,n))^d}$.
Repeating this procedure for each $\dist \in \dists$, with probability at least $ 1 - \numdists \delta$, we have an $\epsilon$-covering $\hyps'$ of $\dists$ of size $\setsize{\hyps'} \in \bigO{\numdists (\mathrm{poly}(1/\epsilon, d,n))^d}$.
We can then appeal directly to Theorem \ref{thm:finite} for a sample complexity bound on learning $(\hyps', \dists)$.
\end{proof}

\begin{restatable}[Assumption 3]{corollary}{marginal}
    \label{corollary:marginal}
  There is an algorithm that, given  access to the marginal distribution $\dist_\features$ of every $\dist \in \dists$, with probability $1-\delta$, returns an  $\epsilon$-optimal solution $\overline{\hyp} \in \simplex(\hyps)$  
    using a number of samples that is
$
	    \bigO{
	    \frac{d\log(dn/\epsilon) + \numdists \log( \numdists / \delta)}{\epsilon^2}
	    }.
 $
\end{restatable}
\begin{proof}
By uniform convergence, taking $\Theta \para{\frac{d \log(d / \epsilon) + \log(1/\delta)}{\epsilon^2}}$ i.i.d.\ samples from distribution $\dist_\features$ for each $\dist \in \dists$, with probability at least $1 - \delta$, yields an $\epsilon$-covering on every $\dist \in \dists$.
By the Sauer-Shelah lemma, the resulting covering $\hyps'_{\dist}$ is of size $\bigO{(\numdists\epsilon^{-2}(\log(d/\epsilon) +  \frac{1}{d} \log(1/\delta) ))^d}$.
We then appeal to Corollary \ref{corollary:covering}.
\end{proof}
\end{proof}

	One question left open by these results is whether, for agnostic collaborative learning, it is possible to achieve sample complexity rates of $\bigO{\epsilon^{-2} \para{\log(n) \vcdt(\hyps) + n \log(n / \delta)}}$ without any additional assumptions or a priori knowledge of an $\epsilon$-net.
    It also remains an open question whether the $\log(n)$ factor in the $\log(n) \vcdt(\hyps) / \epsilon^2$ term is necessary for VC classes, as Theorem~\ref{thm:finite} proves that, for finite/online-learnable classes with sample complexities expressed in terms of $\log(\setsize{\hyps})$ or Littlestone dimension $\lsdt(\hyps)$, no such $\log(n)$ factor is necessary.
    We refer interested readers to \citet{haghtalabcolt} for a complete discussion of these open problems.

\section{Empirical Analysis of On-Demand Sampling for Group DRO}
\label{section:experiments}

This section describes experiments where we adapt our on-demand sampling-based multi-distribution learning algorithm for deep learning applications.
In particular, we compare our algorithm against the de facto standard multi-distribution learning algorithm for deep learning, Group DRO (GDRO)~\cite{sagawa_distributionally_2020}.
As GDRO is designed for use with offline-collected datasets, to provide a meaningful comparison, we modify our algorithm to work on offline datasets (i.e., with no on-demand sample access).

\newcommand{\best}[1]{\textcolor{blue}{\textbf{#1}}}

\begin{table}[ht]
	\small
	\begin{center}
		\begin{tabular}{ c c |c | c | c |c| c| c|}
			\cline{3-8}
			                                          &                  &
			\multicolumn{3}{c|}{Worst-Group Accuracy} &
			\multicolumn{3}{c|}{Gap in Avg. vs. Worst-Group Acc.}                                              \\
			\cline{3-8}
			                                          &                  &
			ERM
			                                          &
			GDRO
			                                          & R-MDL
			                                          &
			ERM
			                                          &
			GDRO
			                                          & R-MDL
			\\
			\cline{2-8}
			\parbox[t]{2mm}{\multirow{3}{*}{\rotatebox[origin=c]{90}{Standard Reg.}}}
			                                          &
			\multicolumn{1}{|l|}{Waterbirds}
			                                          &
			60.0 (1.9)
			                                          &
			76.9 (1.7)
			                                          &
			\best{ 86.4 (1.4)}
			                                          & 37.3 (1.9)
			                                          & 20.5 (1.7)
			                                          & \best{8.1 (1.4)}
			\\
			\cline{2-8}
			                                          &
			\multicolumn{1}{|l|}{CelebA}
			                                          &
			41.1 (3.7)
			                                          &
			41.7 (3.7)
			                                          &
			\best{88.9 (2.3)}
			                                          & 53.7 (3.7)       & 53 (3.7)         & \best{3.4 (2.3)}
			\\
			\cline{2-8}
			                                          &
			\multicolumn{1}{|l|}{MultiNLI}
			                                          &
			66.3 (1.6)
			                                          &
			66.6 (1.6)
			                                          &
			\best{70.3 (1.5)}
			                                          & 16.2 (1.6)       & 15.6 (1.6)       & \best{4.5 (1.5)}
			\\
			\cline{2-8}
			\vspace{-4mm}                                                                                      \\
			\cline{2-8}
			\parbox[t]{2mm}{\multirow{2}{*}{\rotatebox[origin=c]{90}{Strong Reg.}}}
			                                          &
			\multicolumn{1}{|l|}{Waterbirds}
			                                          &
			21.3 (1.6)
			                                          &
			84.6 (1.4)
			                                          &
			\best{89.4 (1.2)}
			                                          & 74.4 (1.6)       & 12 (1.4)         & \best{0.4 (1.3)}
			\\
			\cline{2-8}
			                                          &
			\multicolumn{1}{|l|}{CelebA}
			                                          &
			37.8 (3.6)
			                                          &
			86.7 (2.5)
			                                          &
			\best{88.8 (2.3)}
			                                          & 58 (3.6)         & 6.8 (2.5)        & \best{1.2 (2.3)}
			\\
			\cline{2-8}
			\vspace{-4mm}                                                                                      \\
			\cline{2-8}
			\parbox[t]{2mm}{\multirow{3}{*}{\rotatebox[origin=c]{90}{Early Stop}}}
			                                          &
			\multicolumn{1}{|l|}{Waterbirds}
			                                          &
			6.7 (1.0)
			                                          &
			85.8 (1.4)
			                                          &
			\best{87.1 (1.3)}
			                                          & 87.1 (1.0)       & 7.4 (1.4)        & \best{5.6 (1.3)}
			\\
			\cline{2-8}
			                                          &
			\multicolumn{1}{|l|}{CelebA}
			                                          &
			25.0 (3.2)
			                                          &
			88.3 (2.4)
			                                          &
			\best{90.6 (2.2)}
			                                          & 69.6 (3.2)       & 3.5 (2.4)        & \best{0.7 (2.2)}
			\\
			\cline{2-8}
			                                          &
			\multicolumn{1}{|l|}{MultiNLI}
			                                          &
			66.0 (1.6)
			                                          &
			\best{77.7 (1.4)}
			                                          &
			43.1 (1.7)
			                                          & 16.8 (1.6)       & \best{3.7 (1.4)} & 18.3 (1.7)
			\\
			\cline{2-8}
		\end{tabular}
	\end{center}
	\caption{\small Worst-group accuracy (our primary performance metric) and the gap between worst-group accuracy and average accuracy, of empirical risk minimization (ERM), Group DRO (GDRO), and our R-MDL algorithm in three experiment settings---standard hyperparameters (Standard Reg.), inflated weight decay regularization (Strong Reg.), and early stopping (Early Stop)---and on three datasets---Waterbirds, CelebA, and MultiNLI.
		Figures are percentages evaluated on the test split of each dataset, with standard deviation in parentheses.
		R-MDL consistently outperforms GDRO and performs reliably with or without strong regularization.
	}
	\label{tab:results}
\end{table}

\paragraph{Resampling Multi-Distribution Learning (R-MDL).}
To be more suitable for deep learning applications, we instantiate Algorithm~\ref{alg:collab_simple} by choosing a minibatch gradient descent algorithm as the minimizing player's algorithm ($\learningalgorithm_{\Theta}$) and a naive uniform-sampling bandit algorithm as the maximizing player's algorithm ($\learningalgorithm_{\simplex(\dists)}$).
We can further adapt our algorithm to offline datasets by simulating on-demand sampling on the empirical distributions of datasets.
This modified algorithm, R-MDL, is described in full in Algorithm~\ref{alg:collab_deep}.

Note that, in contrast, the original group DRO algorithm of \citet{sagawa_distributionally_2020} is also a minibatch gradient descent algorithm but samples minibatches uniformly from all distributions and weights datapoints via a no-regret algorithm that provides importance weights.
Though effective, this method is brittle and requires tricks like unconventionally strong regularization~\cite{sagawa_distributionally_2020}.
Our theory of on-demand sampling suggests that R-MDL should mollify this brittleness, as it replaces GDRO's upweighting of low-accuracy distributions with upsampling of low-accuracy distributions.
Interestingly, the advantage of \textit{resampling} over \textit{reweighting} has been previously observed when training neural networks on a dataset with fixed importance weights~\cite{sagawa_investigation_2020}.

\paragraph{Experiment Setting}
In Table \ref{tab:results}, we replicate the Group DRO experiments of~\citet{sagawa_distributionally_2020} and compare the standard GDRO algorithm with our R-MDL algorithm (Algorithm~\ref{alg:collab_deep}).
We fine-tune Resnet-50 models (convolutional neural networks)~\cite{he_deep_2016} and BERT models (transformer-based network)~\cite{devlin_bert_2019} on the image classification datasets Waterbirds~\cite{sagawa_distributionally_2020,wah_caltech-ucsd_2011} and CelebA~\cite{liu_deep_2015} and the natural language dataset MultiNLI~\cite{williams_broad-coverage_2018} respectively.
We train these models in 3 settings: with standard hyperpameters, under strong weight decay ($\loss$-2) regularization, or under early stopping.

\paragraph{R-MDL consistently outperforms GDRO and ERM.}
In every dataset and in almost every setting, R-MDL significantly outperforms GDRO and ERM in worst-group accuracy. %
In addition, whereas GDRO and ERM have large gaps between worst-group accuracy and average accuracy, R-MDL has almost matching worst-group and average accuracies.
This indicates that R-MDL is more effective at prioritizing learning on difficult groups.

\paragraph{R-MDL is robust to regularization strength.} R-MDL retains high worst-group accuracy even without strong regularization.
These results challenge the observation of~\citet{sagawa_distributionally_2020} that strong regularization is critical for the performance of Group DRO methods.
This suggests that the brittleness of GDRO is due to the reweighting rendering the adversary too weak.
In contrast, R-MDL provides a robust multi-distribution learning method with significantly less hyperparameter sensitivity. %

\section{Conclusions}
While learning from a single data distribution is a fundamental abstraction of data-driven pattern recognition, data-driven decision-making calls for a new perspective that captures learning problems involving multiple stakeholders and data sources.
This work proposes multi-distribution learning as a unifying theoretical framework, bringing together a number of widely studied problem formulations such as group distributionally robust optimization and collaborative PAC learning under a single umbrella.
This unifying perspective distills the challenges of these various learning problems to a fundamental question about the sample complexity of stochastic games.
We answered this fundamental question by providing optimal rates for a broad class of problems including convex and Littlestone hypothesis classes, highlighting the importance of on-demand sampling for decoupling the complexity of learning and obtaining robustness.
We believe these findings underscore a broader takeaway that adaptive data collection is fundamental for scalable learning outside the single-distribution paradigm of classical pattern recognition.

\section{Acknowledgments}
This work was supported in part by the National Science Foundation under grant CCF-2145898, a C3.AI Digital Transformation Institute grant, and the Mathematical Data Science program of the Office of Naval Research.
This work was partially done while Haghtalab and Zhao were visitors at the Simons Institute for the Theory of Computing.
The authors thank the authors of \citet{chen_tight_2018} and \citet{sagawa_distributionally_2020} for communication regarding their work.
The authors also thank Tianyi Lin, Guy Rothblum, Abhishek Shetty, Tatjana Chavdarova, Lydia Zakynthinou, and Mingda Qiao for valuable discussions.

\newpage
\bibliographystyle{abbrvnat.bst}
% \bibliography{lifelong_out}
\newcommand{\nips}[1]{Advances in Neural Information Processing Systems #1}

\newpage
\appendix

\section{Experiment Details}
\def\dataset{\mathcal{X}}
\begin{algorithm}[t]
	\caption{Resampling-based Multi-Distribution Learning (R-MDL)}
	\label{alg:collab_deep}
	\begin{algorithmic}
		\STATE \textbf{Input:} Parameter space $\Theta$, iterations $T$, batch size $B$ and adversary batch size $B'$, and training and validation datasets $X_{\text{tr}, i}$ and $X_{\text{val}, i}$ for $i \in [\numdists]$;
		\STATE \textbf{Initialize:} $\tsv{\theta}{0} \in \Theta$ and $\tsv{w}{0} = [1/n]^n$;
		\FOR{$\timestep = 1, 2, \dots, T$}
		\STATE For $i \in [\numdists]$, randomly sample (with replacement) $B'$ datapoints $\tsv{x}{t-1}_{\text{val}, i, 1}, \dots, \tsv{x}{t-1}_{\text{val}, i, B'}$ from $X_{\text{val}, i}$;
            \STATE Let $\tsv{w}{t} = \hedgealgorithm_{\simplex_\numdists}\para{\bset{w \mapsto 1-\frac{1}{B'}\sum_{j=1}^{B'} \sum_{i=1}^\numdists w_i \loss(\hyp_{\tsv{\theta}{\tau}}, \tsv{x}{\tau}_{\text{val}, i, j})}_{\tau \in [t-1]}}$, see \equationref{\ref{eq:proxmapping}};
            \STATE Randomly sample (with replacement) the datapoints $\tsv{x}{t-1}_{\text{tr}, 1}, \dots, \tsv{x}{t-1}_{\text{tr}, B}$ from $\sum_{i=1}^\numdists \tsv{w}{t-1}_i \dist_i$;
		\STATE Run a gradient descent update(s) $\tsv{\theta}{t} = \mathrm{GradientDescent}_\Theta\para{\theta \mapsto \frac{1}{B}\sum_{j=1}^{B} \loss(\hyp_\theta, \tsv{x}{\tau}_{\text{tr}, j})}_{\tau \in [t-1]}$;
		\ENDFOR
		\STATE \textbf{Return:} $\frac 1T \sum_{t=1}^T \tsv{\theta}{t}$;
	\end{algorithmic}
\end{algorithm}

\paragraph{R-MDL Algorithm.}
The R-MDL algorithm is defined in full in Algorithm~\ref{alg:collab_deep}.
It instantiates (a batched version of) Algorithm~\ref{alg:collab_simple} choosing $\learningalgorithm_\Theta$ to be online gradient descent and $\learningalgorithm_{\simplex_\numdists}$ to be a naive bandit-to-full-information reduction algorithm that implements Exp3 but observes cost functions uniformly at random and re-uses cost function observations between rounds.
This algorithm is an example of instantiating our general multi-distribution learning framework with more practical choices of learning algorithms.
An example implementation, along with experiment replications, is provided in the Github repository \href{https://github.com/ericzhao28/multidistributionlearning}{ericzhao28/multidistributionlearning}.

\begin{figure}
	\centering
	\includegraphics[width=0.6\textwidth]{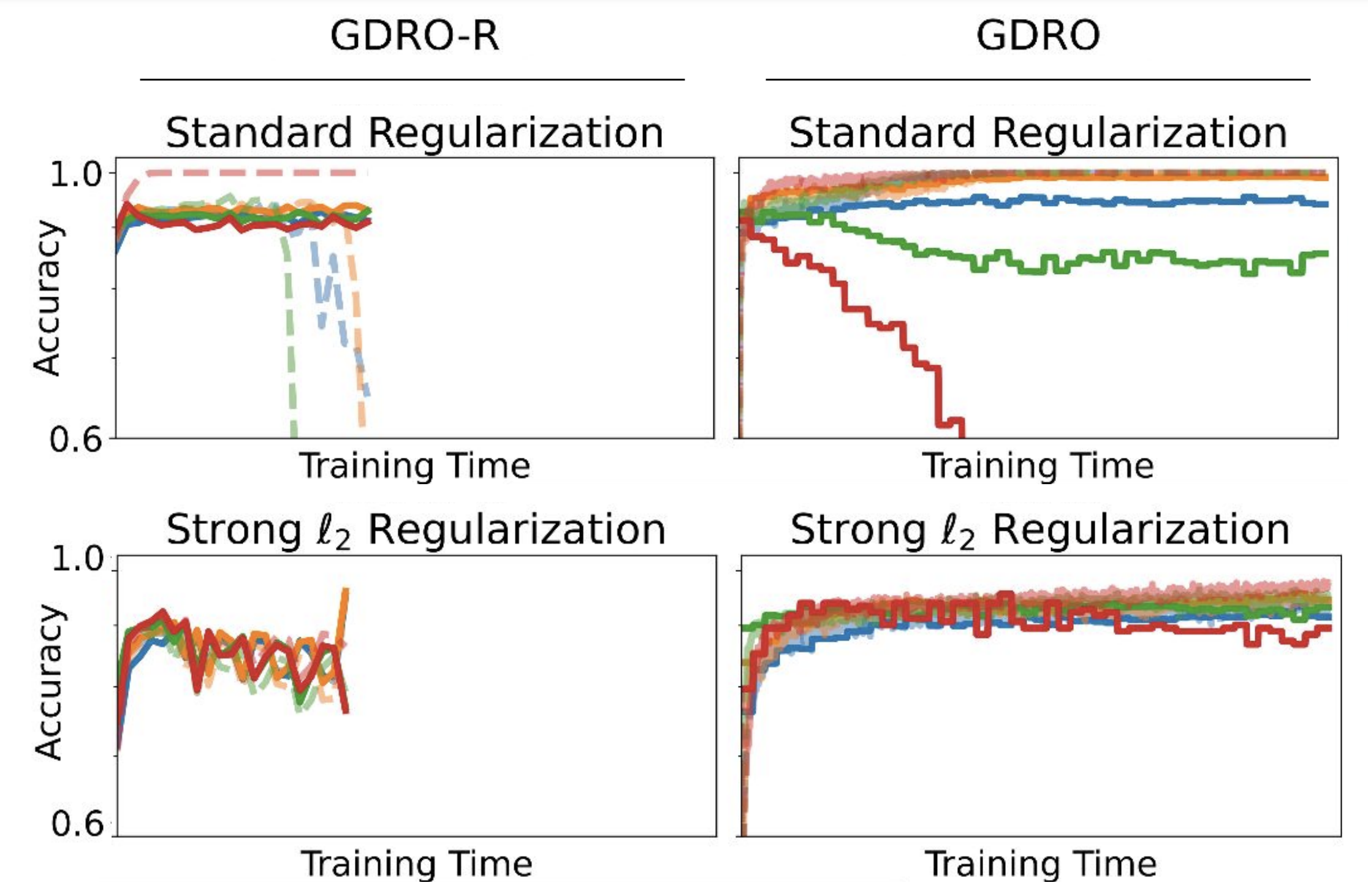}
	\caption{
		Training (light, dashed) and validation (dark, solid) accuracies for GDRO and R-MDL during training, plotted on a log scale.
		Note that R-MDL validation accuracy will be noisier than those of GDRO as we constrain R-MDL to limited samples (with replacement) from the validation set.
		In addition, in the left-most plot, training accuracy for all groups except the blond male group (red) dips to zero due to lack of data---this is because the blond male group (red) is the most challenging so the adversary eventually stops sampling from other groups.
		Under standard regularization, the red-group accuracy drops off in GDRO while R-MDL maintains a high red-group accuracy by heavily sampling from the red group, as reflected in the near-perfect red-group training error.
	}
	\label{fig:graph}
\end{figure}

\paragraph{Additional Observation: R-MDL converges faster than ERM or GDRO.}
The R-MDL methods in Table \ref{tab:results} used a fraction of the training epochs that their GDRO counterparts used.
The ratio of R-MDL to GDRO training epochs is 1:3, 2:5, 1:2 on the Waterbirds, CelebA, and MultiNLI datasets respectively.
This fast convergence rate is predicted by our theory, particularly Corollary \ref{corr:gdro}.
In our Figure \ref{fig:graph}, we also replicate the Figure 2 of~\citet{sagawa_distributionally_2020}, appending our additional results on R-MDL.
We again see a trend of faster test error convergence (solid lines) and more uniform per-group risks by the R-MDL algorithm.

\paragraph{Datasets.}
Our experiments were performed on three datasets: Multi-NLI, CelebA, and Waterbirds~\cite{sagawa_distributionally_2020}.
We use identical preprocessing settings and dataset splits as~\citet{sagawa_distributionally_2020}.
Our experiments, unless otherwise specified, replicate the exact hyperparameter settings adopted by~\citet{sagawa_distributionally_2020} for their Table 2 experiments.
This includes the choice of random seeds, batch sizes, learning rates, learning schedules, and regularization.
We defer readers to~\citet{sagawa_distributionally_2020} or to our public source code for replication details.

The \textbf{Multi-NLI dataset}~\cite{williams_broad-coverage_2018} concerns the following natural language inference task: determine if one statement is entailed by, neutral with, or contradicts a given statement.
This dataset is challenging because traditional ERM models are prone to spuriously correlating ``contradiction'' labels with the existence of negation words.
The dataset is divided into 6 distributions: the Cartesian product of the label space (entailment, neutral, contradiction) and an indicator of whether the sentence contains a negation word.
The label space annotations were annotated by~\cite{williams_broad-coverage_2018} while negation labels were annotated by~\citet{sagawa_distributionally_2020}.
There are 206,175 datapoints available in the Multi-NLI dataset; the smallest distribution (entailment + negation) is represented by only 1,521 datapoints.
We use a randomly shuffled 50-20-30  training-validation-testing split.

The \textbf{CelebA dataset} is a dataset of celebrity face images and a label space of potential physical attributes~\cite{liu_deep_2015}.
This dataset is challenging because traditional ERM models are prone to spuriously correlating attribute labels with demographic information such as race and gender.
Following~\citet{sagawa_distributionally_2020}, we divide the dataset into 4 distributions: the Cartesian product of the blond vs dark hair attribute label (``Blond\_Hair'') with the ``gender'' attribute label (``Male'').
Note that the authors of~\citet{liu_deep_2015} limited the ``gender'' attribute label to binary options of male and not male.
There are 162,770 datapoints available in the CelebA dataset; the smallest distribution (blond-hair + male) is represented by only 1,387 datapoints.
We use the official training-testing-validation dataset split.

The \textbf{Waterbirds dataset} is a dataset by~\citet{sagawa_distributionally_2020} curated from a larger Caltech-UCSD Birds-200-2011 (CUB) dataset~\cite{wah_caltech-ucsd_2011}.
It concerns the task of predicting whether a bird is of some waterbird (sub)species from an image of said bird.
This dataset is challenging because traditional ERM models are prone to spuriously correlating backgrounds with foreground subjects; for instance, a model may often predict that a bird is a waterbird only because the image of the bird was taken at a beach.
The dataset has 4 distributions: the Cartesian product of the waterbird vs not waterbird label with whether the background of the picture is over water.
There are 4,795 datapoints available in the Waterbirds dataset; the smallest distribution (waterbirds on land) is represented by only 56 examples.

\paragraph{Models.}
We use two classes of models in our experiments: Resnet-50~\cite{he_deep_2016} and BERT~\cite{devlin_bert_2019}.
We use the \textit{torchvision}~\cite{marcel_torchvision_2010} implementation of the convolutional neural network Resnet-50, with a default choice of a stochastic gradient descent optimizer with momentum 0.9 and batch size 128. Batch normalization is used; data augmentation and dropout are not used.
We use the \textit{HuggingFace}~\cite{wolf_huggingfaces_2019} implementation of the language model BERT, with a default choice of an Adam optimizer with dropout and batch size 32.

\paragraph{Hyperparameters.}
In the \textit{Standard Regularization} experiments, we use a Resnet-50 model with an $\loss$-2 regularization parameter of $\lambda=0.0001$ and a fixed learning rate of $\alpha = 0.001$ for both Waterbirds and CelebA datasets.
The ERM and Group DRO baselines are trained on CelebA for 50 epochs and Waterbirds for 300 epochs.
Our multi-distribution learning method is trained on CelebA for only 20 epochs and Waterbirds for 100 epochs; this is due to the faster training error convergence of our method.
For the MultiNLI dataset, we use a BERT model with a linearly decaying learning rate starting at $\alpha_0 = 0.00002$ and no $\loss$-2 regularization.
The ERM and Group DRO baselines are trained on Multi-NLI for 20 epochs.
Our multi-distribution learning method is trained on Multi-NLI for only 10 epochs.
Our multi-distribution learning method uses adversary learning rates $\maxplayerx{\eta}$ of 1, 1, 0.2 on Waterbirds, CelebA and MultiNLI respectively.

In the \textit{Strong Regularization} experiments, we follow similar settings to the \textit{Standard Regularization} experiments.
The only change is that an $\loss$-2 regularization parameter of $\lambda=1$ is used for Waterbirds and an $\loss$-2 regularization parameter of $\lambda =0.1$ is used for CelebA.
Our multi-distribution learning method uses adversary learning rates $\maxplayerx{\eta}$ of 1 and 0.2 on Waterbirds and CelebA respectively.

In the \textit{Early Stopping} experiments, we follow similar settings to the \textit{Standard Regularization} experiments.
The only change is that all CelebA and Waterbird experiments are run for a single epoch.
MultiNLI experiments are run for 3 epochs.
Our multi-distribution learning method uses adversary learning rates $\maxplayerx{\eta}$ of 1, 1, 1 on Waterbirds, CelebA and MultiNLI respectively.

The only hyperparameters we use that differ from prior literature are the number of training epochs and the adversary learning rates of our method (R-MDL).
The choice of epoch was not fine-tuned, and was selected due to our observation of early training error convergence.
We selected our adversary learning rate $\minplayerx{\eta}$ by training our method, on each dataset, for both $\minplayerx{\eta}=1$ and $\minplayerx{\eta}=0.2$ and selecting the $\minplayerx{\eta}$ yielding the highest validation-split worst-group accuracy.

\paragraph{Compute.}
The total amount of compute run for the experiments in this section is approximately 50 GPU hours.
A ``n1-standard-8'' machine was leased from the Cloud computing service Google Cloud; the ``n1-standard-8'' machine was equipped with 8 Intel Broadwell chips and 1 NVIDIA Tesla V100 GPU.
The cost of these computing resources totaled approximately USD \$2 per hour, with a total cost of approximately USD \$100.
All results described in this section, with the exception of existing results cited from other works, were obtained with experiments on said machine.
All experiments were implemented in Python and PyTorch.

\end{document}